\UseRawInputEncoding

\documentclass{article}

\usepackage{microtype}
\usepackage{graphicx}
\usepackage{subcaption}
\usepackage{booktabs} 
\usepackage{multirow}
\usepackage{tabularx}
\usepackage{paralist}
\usepackage{wrapfig}
\usepackage[most]{tcolorbox}

\usepackage{hyperref}

\usepackage[preprint]{neurips_2026}

\usepackage{amsmath}
\usepackage{amssymb}
\usepackage{mathtools}
\usepackage{amsthm}
\usepackage{bbm}

\newcommand{\ppt}{pp_{\text{target}}}

\usepackage[capitalize,noabbrev]{cleveref}

\theoremstyle{plain}

\theoremstyle{definition}

\theoremstyle{remark}

\usepackage{authblk}
\usepackage[textsize=tiny]{todonotes}

\title{Predicting Performance of Symbolic and Prompt Programs with Examples}
\author[1,*]{Chengqi Zheng}
\author[2,*]{Keya Hu}
\author[1]{Shuzhi Liu}
\author[1]{Tao Wu}
\author[3]{Kevin Ellis}
\author[1,$\dagger$]{Yewen Pu}
\affil[1]{Nanayang Technological University, Singapore}
\affil[2]{Massachusetts Institute of Technology, USA}
\affil[3]{Cornell University, USA}
\affil[$\dagger$]{Corresponding Author}

\begin{document}

\maketitle
\vspace{-20pt}
\begin{abstract}
LLM prompting is widely used for naturally stated tasks, yet it is unreliable --- it may succeed on a few test cases but fail at deployment time.
We study \textit{performance prediction}: given a program --- either symbolic (e.g. Python) or a prompt executed on an LLM, and a few in-domain examples, predict its performance on unseen tasks from the same domain.
We use a simple coin-flip model, treating each pass/fail program execution as a Bernoulli random variable, whose success probability $\theta$ is the program’s unknown performance.
In this model, performance depends entirely on: 1) the observed execution outcomes on test cases, and 2) a prior over performances.
We compile \textit{empirical performance priors} from a corpus of diverse programs and tasks, and find that performance for symbolic programs (e.g., Python) are “all or nothing,” while prompt programs have a diffuse prior with many nearly-correct programs.
This difference explains why a few passing tests can certify symbolic programs but not prompt programs.
Building on this insight, we develop RAP (Retrieved Approximate Prior), which retrieves similar tasks and prompt programs from an existing corpus to construct a proxy prior, which is then used to predict performance. We show RAP achieves solid performances.

\end{abstract}

\newcommand{\SP}{$sp$}
\newcommand{\PP}{$pp$}

\section{Introduction}
\label{introduction}

If a program passes a few representative test cases, one can reasonably trust it to remain correct on future instances of the same task. 
The same is not true for LLM prompting—prompts may appear correct during development yet fail unexpectedly at deployment \citep{zhou2025predictableartificialintelligence}. Why do a few examples certify symbolic programs (e.g., Python code) but not prompt programs (natural-language instructions executed by an LLM)? We study this question through a Bayesian lens, modeling test outcomes as independent coin tosses and using them to infer a program’s performance.

Prior work shows that prompting is brittle \citep{mizrahi2024stateartmultipromptllm} --- small wording or style changes can cause large performance drops, and that even strong LLMs can fail on simple tasks \citep{Zhou2024LessReliable}. 
Yet prompting remains widely used because it is general, simple, and often “good enough” for non-critical settings where a performance of 80\% is still valuable. 
This motivates \emph{performance prediction}: given a prompt program and a few test cases, infer a distribution over its true success rate so one can decide whether a prompt is deployable and with what confidence (e.g., “likely $\sim 0.8 \pm 0.1$ once deployed”).

\begin{wrapfigure}{r}{0.62\columnwidth}

\centering
\includegraphics[width=0.60\columnwidth]{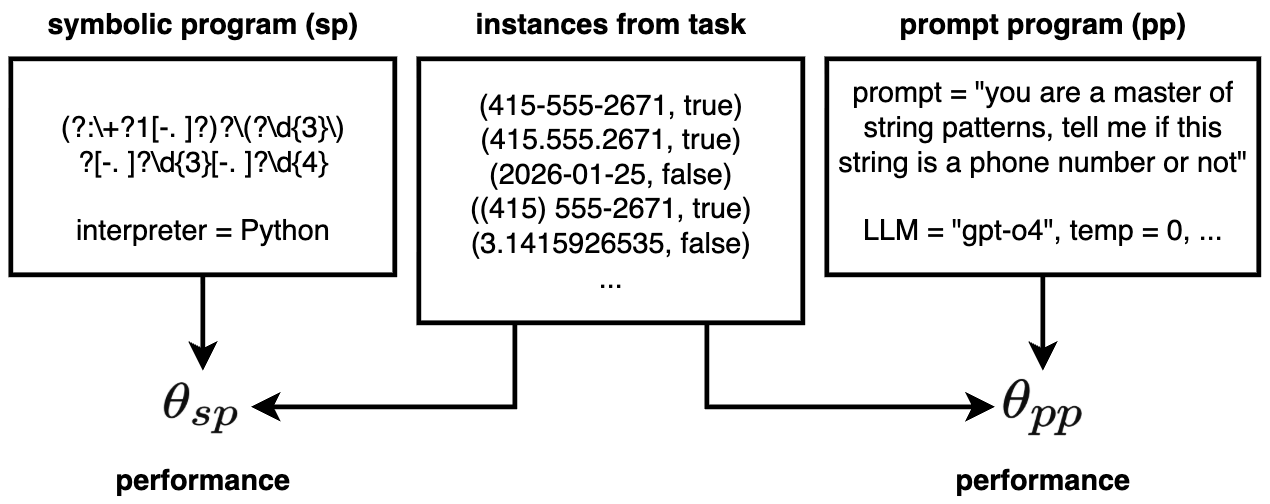}

\caption{Two ways of solving a phone-number validation task. A symbolic program (\SP{}) uses an explicit pattern, whereas a prompt program (\PP{}) delegates prediction to an LLM. In both cases, performance $\theta$ is the fraction of task instances solved correctly. Note we consider the choice of LLM and parameters such temperature \emph{as a part of} the prompt program.}
\label{fig:sp_pp}
\vspace{-1.0em}
\end{wrapfigure}

This work is the first to study performance prediction for both \emph{symbolic programs} (\SP{}) and \emph{prompt programs} (\PP{}). 
We define a symbolic program as one whose deployed behavior is executed by a symbolic interpreter (e.g., Python, regex, or another formal language), whereas a prompt program is executed by an LLM at deployment time. 
The key distinction is \textit{what is delegated at deployment}: if an LLM is used only to \emph{write} Python code and the deployed artifact is that code, we categorize it as \SP{}; in contrast, if the deployed artifact is a natural-language prompt that must be run by an LLM to produce outputs for new inputs at deployment, we categorize it as \PP{}. 
\Cref{fig:sp_pp} illustrates a pattern-matching task that can be solved as either \SP{} or \PP{}, and in both cases we measure performance $\theta$ by evaluating the solution on instances sampled from the task.

We adopt a simple performance model that treats each evaluation outcome as a Bernoulli coin toss.
Under this model, given an observation summarized by $O=(a,b)$ (with $a$ passes and $b$ fails on some test-cases), the posterior distribution over performance is $P(\theta|O) \propto P(\theta)P(O|\theta)$.
While the likelihood $P(O|\theta) \propto \theta^a(1-\theta)^b$ can be easily computed, the prior of performance $P(\theta)$ remains unknown.
To resolve this, we compiled empirical performance priors for \SP{} and \PP{} from a diverse set of tasks and programs.
These empirical priors reveal a stark difference: \SP{} exhibits an ``all-or-nothing'' prior concentrated near 0.0 and 1.0, whereas \PP{} has a more diffuse prior with substantial mass on intermediate, nearly-correct performance (\Cref{fig:sp-vs-pp-prior}).
Consequently, passing a few test-cases can certify a \SP{}'s performance, while leaving the performance of \PP{} uncertain (\Cref{fig:taifeng}).
Thus, our simple performance model can \textit{give quantification} to the intuition that symbolic code tends to be well certified by few test cases, while prompt programs do not.

Building on this insight, we posit that in order to accurately predict performance of \PP{}s, we need to construct a better prior.
We develop \textbf{RAP} (\emph{Retrieval Approximated Prior}).
Given a prompt program, a few in-domain test cases, and an \emph{arbitrary} corpus of tasks and prompts, RAP retrieves similar tasks and prompt programs from the corpus, and constructs an approximate prior for performance prediction.
We show RAP outperforms baseline performance predictors and has several desirable properties: it converges to the in-domain posterior as the number of test cases grows, improves with larger corpora, and is robust to irrelevant information in the corpus.

\noindent\textbf{Contributions~}
We make three equally weighted contributions:
\begin{compactenum}
    \item We formalize \emph{performance prediction} as Bayesian inference under a simple coin-flip model, inferring a program's performance from observed execution outcomes $P(\theta \mid O)$.
    \item We compile empirical priors for \SP{} and \PP{} from a diverse set of tasks and programs, and show that \SP{} is ``all-or-nothing,'' while \PP{} is diffuse and nearly-correct-heavy. This explains why a few tests can certify \SP{} but not \PP{}.
    \item We introduce \textbf{RAP} (\emph{Retrieval Approximated Prior}), a retrieval-based method that builds an approximate domain-specific prior from a corpus and outperforms strong baselines.
\end{compactenum}

\newcommand{\Perf}{\theta_f}
\newcommand{\Z}{\mathbf{Z}}
\section{Predicting Performance from Examples}
\label{sec:predict_performance}


This section formalizes the problem of \emph{performance prediction from 
examples}. We first define what we mean by a program and its performance on a
task, and then define the objective of predicting such performance and
how to evaluate prediction quality.

\subsection{Definitions}
\textbf{tasks}
A task $T$ is a distribution of instances $(x,y) \sim T$ with input $x$ and oracle-labeled output $y$.
For instance, in the task of sorting, an input may be any random array of integers, and the output is the same array of integers ordered from smallest to largest.
In this work, we assume sampling of instances is iid.

\textbf{programs}
A program $f$ is a mapping from input to output $f: x \rightarrow y$.
Given a task $T$, one writes a program by sampling from $f \sim P(f|T)$. 
In practice, one may either draw a sample by asking a person to write a program for a task, or by prompting an LLM to write a program for a task.

\textbf{correctness}
Given an instance $(x,y)$, a program's correctness on this instance is an indicator $\Z = \mathbbm{1}(f(x)=y)$

\textbf{performance}
Given a program $f$ and a task $T$, the function's (true) performance $\theta_f^*$ is the expected success rate of $f$ producing the correct output $y$ on input $x$ for $(x,y) \sim T$
\begin{equation}
\theta_f^*(f, T) = 
\mathbb{E}_{(x,y) \sim T} [ \mathbbm{1}(f(x)=y)]
 .
\end{equation}
In this paper, we simply denote performance with $\theta_f$, implicitly assuming every program is written \emph{for some task}.

\textbf{observation}
An observation is a set of $K$ indicators
\begin{equation}
    \Z_{1 \dots K}, \Z_i = \mathbbm{1}(f(x)=y), (x_i, y_i) \sim T
\end{equation}
We can summarize the outcome of the observation by the number of successes $a$ and the number of failures $b$
\begin{equation}
    O = (a,b), a = \sum_{i} Z_i, b = \sum_i (1-Z_i)
\end{equation}

\setlength{\tabcolsep}{4pt}

\begin{table*}[ht!]
\centering
\small
\setlength{\tabcolsep}{6pt}
\renewcommand{\arraystretch}{1.15}
\caption{Benchmark domains used in this study, covering a diverse range of task structures, input modalities, and symbolic solver representations. This diversity enables a systematic evaluation of performance prediction across heterogeneous program types. Across all domains, we evaluate a total of 700 symbolic programs and 700 prompt programs, with approximately 100 evaluation instances per task.}
\label{tab:sp_vs_pp_tasks}
\begin{tabular}{@{} l l l r r @{}}
\toprule
\textbf{Name} & \textbf{Task given as} & \textbf{Symbolic program type} & \textbf{\#Tasks} & \textbf{\#Progs/task} \\
\midrule
Structured regex~\cite{ye2020benchmarkingmultimodalregexsynthesis} & Natural language & Regex expression & 20 & 10 \\
HumanEval~\cite{chen2021evaluatinglargelanguagemodels}        & Natural language & Python script    & 10 & 10 \\
ARC~\cite{chollet2019measure}              & Few-shot examples & Python script    & 20 & 10 \\
List~\cite{rule2024symbolic}             & Few-shot examples & Lambda function  & 20 & 10 \\
\bottomrule
\end{tabular}

\end{table*}

\subsection{Predicting Performance from Examples}
Given observed outcomes $O$ of running a program $f$ on a set of sampled instances, we wish to infer a program's true performance:
\begin{equation}
    P(\theta_f | O)
\end{equation}

This posterior distribution can be evaluated in two ways:

\textbf{probability density at the ground truth}
We plug in the ground-truth value $\theta_f^*$ into the posterior distribution and obtain its probability density value, $P(\theta_f = \theta_f^* | O)$. 
A poor posterior would have a low score, while a good posterior would have a high score, with the limit approaching infinity (a Dirac-delta posterior).

\textbf{absolute error}
We take the mean of the posterior as the prediction, and take the absolute difference between the prediction value and the ground-truth value $|\hat{\theta_f} - \theta_f^*|$.

\subsection{A Simple Coin-flip Model}
In this work, we take the simplest model of estimating $\theta_f$ by treating each observation $\Z_i$ as a Bernoulli random variable:
\begin{equation}
    \Z_i \sim Bernoulli(\theta_f)
\end{equation}
The likelihood function for observation $O=(a,b)$ follows the binomial distribution (written without the constant):
\begin{equation}
    P(O|\theta_f) \propto \theta_f^a(1-\theta_f)^b
\end{equation}

Thus, the posterior of a program's performance under observation can be written via Bayes' rule: 
\begin{equation}
    P(\theta_f | O) \propto P(\theta_f) \theta_f^a(1-\theta_f)^b
\end{equation}

As we can see, the \textbf{performance prior} $P(\theta_f)$ plays a pivotal role in determining the posterior distribution $P(\theta_f|O)$. 
In the next section we will explain how to empirically estimate the performance prior for both symbolic programs and prompt programs.

\begin{figure}[t]
\centering

\begin{subfigure}[t]{0.48\columnwidth}
\centering
\includegraphics[width=\linewidth]{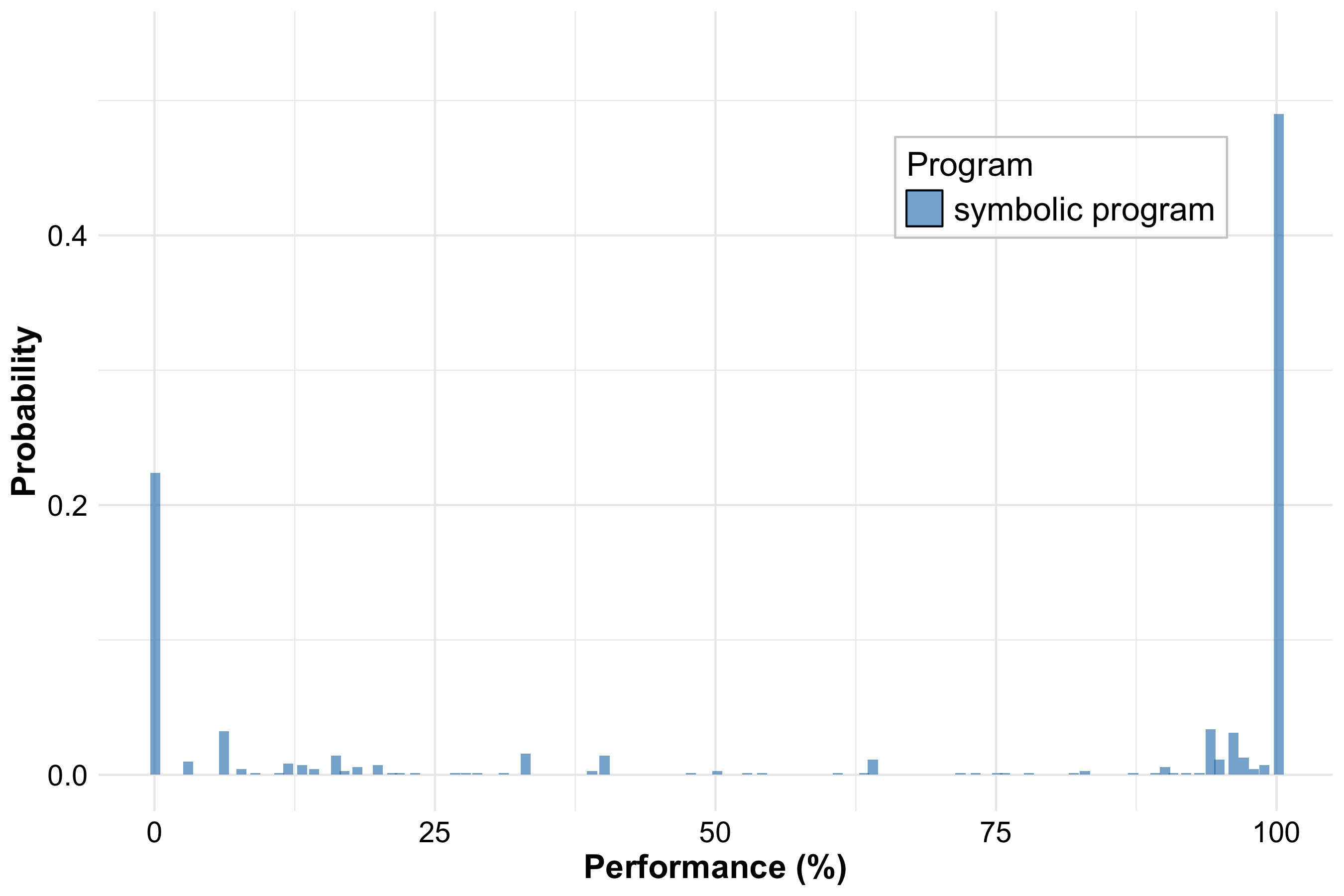}
\caption{Performance prior for symbolic programs}
\label{fig:sp-prior}
\end{subfigure}
\hfill
\begin{subfigure}[t]{0.48\columnwidth}
\centering
\includegraphics[width=\linewidth]{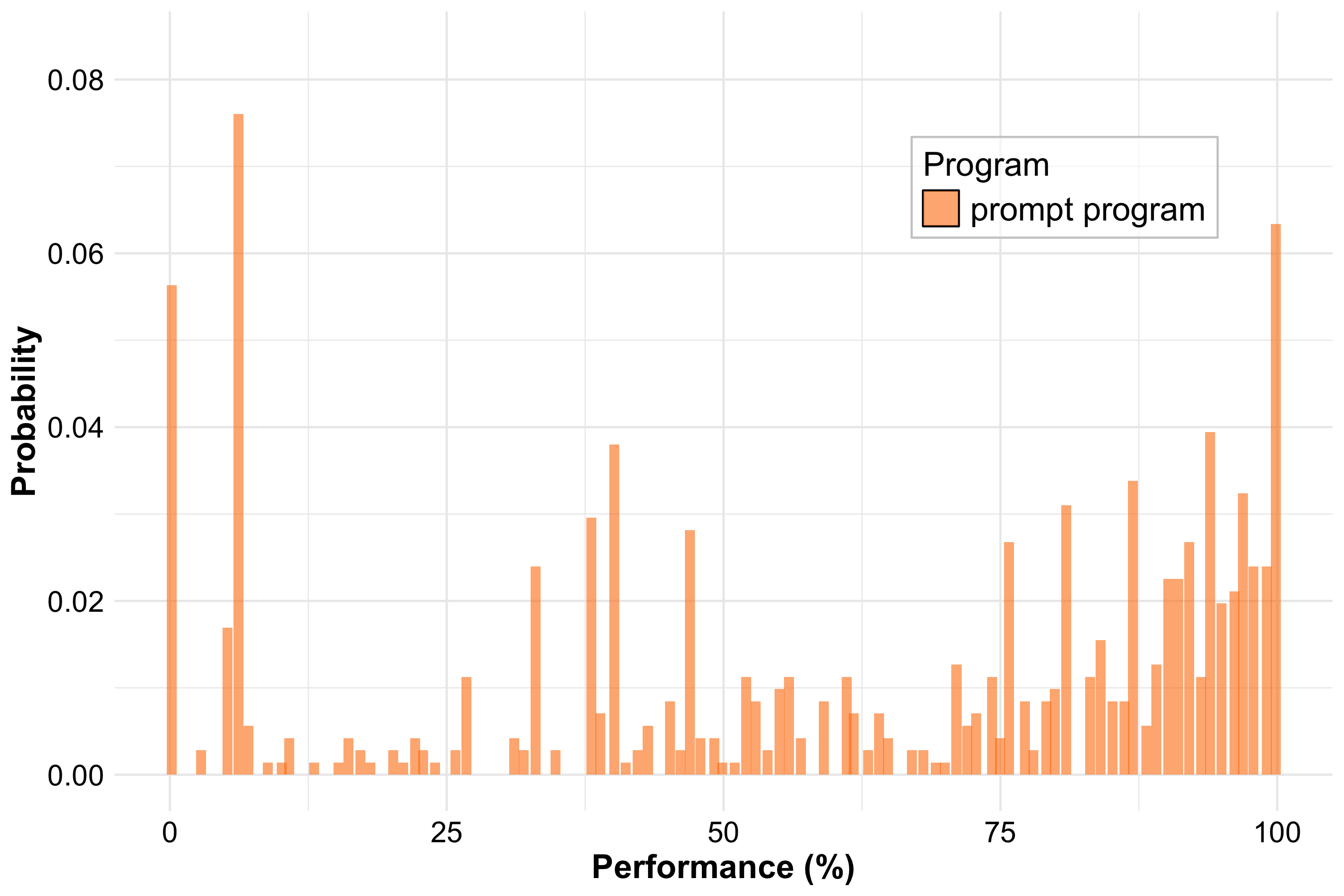}
\caption{Performance prior for prompt programs}
\label{fig:pp-prior}
\end{subfigure}

\caption{
Empirical performance priors for \SP{} and \PP{}, constructed from 700 sampled programs each.
Symbolic programs have a sharply bimodal prior, with most programs either perfect or completely incorrect. Prompt programs have a more diffuse prior, with substantial mass over intermediate performances.
}
\label{fig:sp-vs-pp-prior}
\end{figure}

\begin{figure}[t]
\centering

\begin{subfigure}[t]{0.48\columnwidth}
\centering
\includegraphics[width=\linewidth]{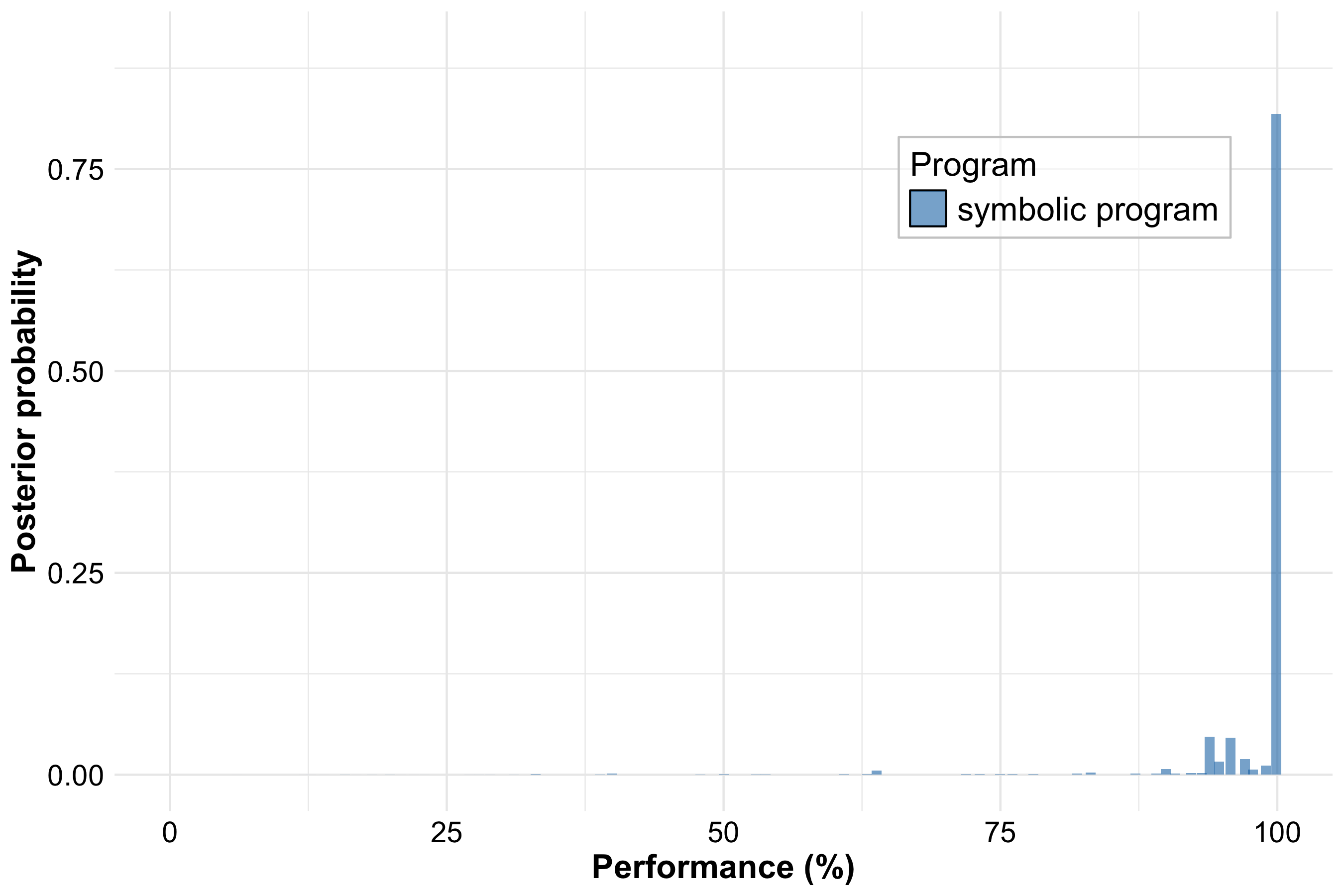}
\caption{Posterior for \SP{} after successes on 3 test cases}
\label{fig:sp-posterior}
\end{subfigure}
\hfill
\begin{subfigure}[t]{0.48\columnwidth}
\centering
\includegraphics[width=\linewidth]{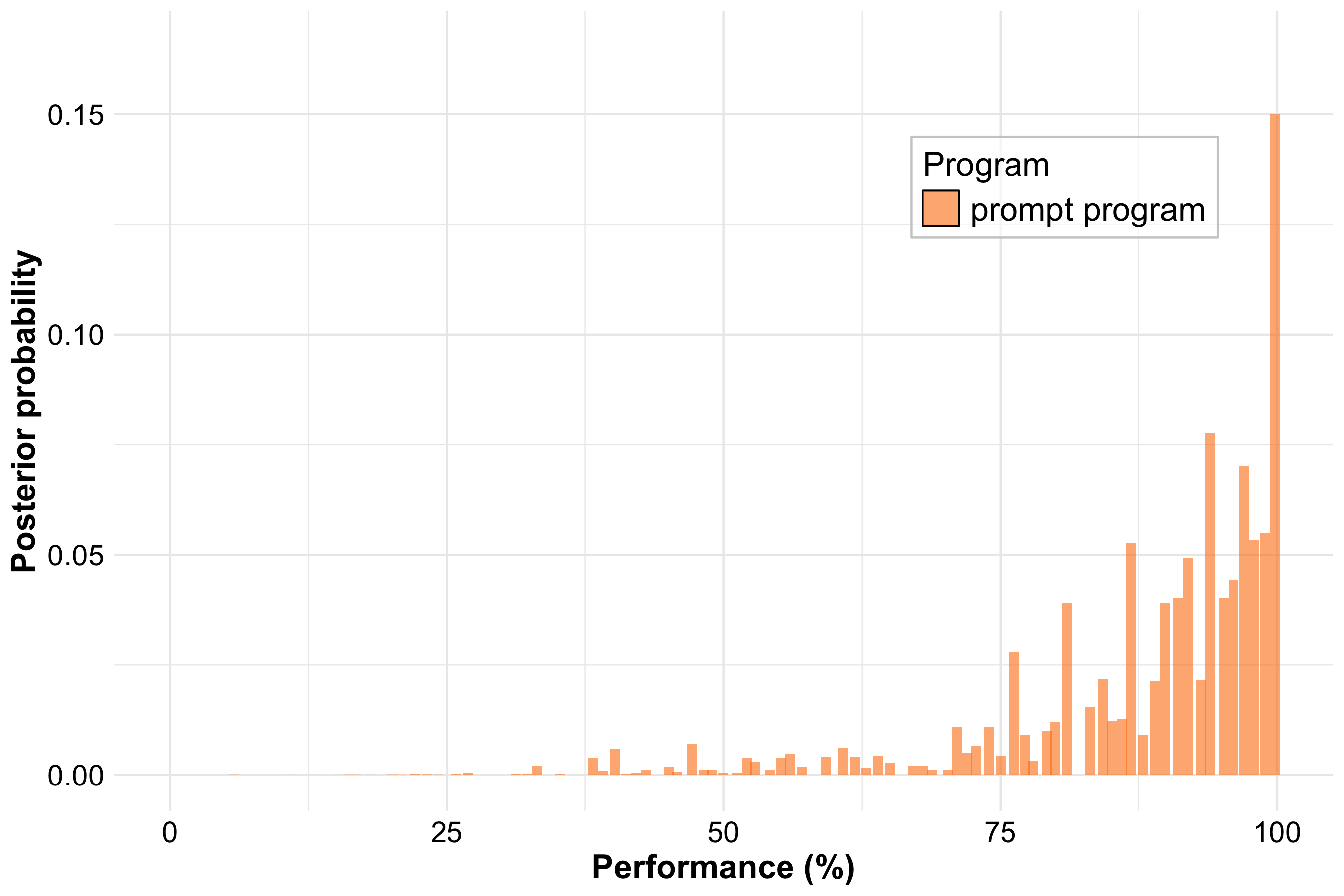}
\caption{Posterior for \PP{} after successes on 3 test cases}
\label{fig:pp-posterior}
\end{subfigure}

\caption{
Posterior performance distributions for symbolic programs $P(\theta_{sp}\mid[T,T,T])$ and prompt programs $P(\theta_{pp}\mid[T,T,T])$ after observing successes on 3 test cases. The likelihood update, $\theta^3$, is the same for both \SP{} and \PP{}; the difference in the posterior is entirely due to their different performance priors. The posterior for \SP{} supports a near-100\% performance prediction, whereas the posterior for \PP{} does not provide a confident performance guarantee.
}
\label{fig:taifeng}
\end{figure}

\section{The Performance Prior}

The performance prior can be written as:
\begin{equation}
P(\theta)
=
\Pr_{T\sim P(T),\, f\sim P(f\mid T)}
\left[
\theta = \theta_f
\right].
\end{equation}
In words, we first sample a task $T \sim P(T)$, and from this task sample a program $f \sim P(f|T)$ (either from a human or from a LLM), then evaluating its performance on the task.
We do this many times, and we would obtain a large list of performance values $[\theta_{f_1}, \theta_{f_2}, \dots]$, which together forms the performance prior.
Ideally, we would like a ``perfect'' performance prior derived from ``all tasks that we have ever written a program for'', which is unrealistic.
Instead, we approximate the performance prior for both \SP{} and \PP{} empirically.

\subsection{Empirical Performance Prior}

\paragraph{sampling tasks $P(T)$}
We chose a diverse set of programming domains --- from inductive reasoning to code generation from language, then randomly sample tasks from these domains.
The domains and numbers of tasks chosen is shown in Table \ref{tab:sp_vs_pp_tasks}.
Crucially, tasks from these domains maybe reasonably solved by both \SP{} and \PP{}. 
For instance, in the Regex domain, one can either solve the task of writing a regular expression (\SP{}) that matches a phone number pattern, or by asking a LLM directly (\PP{}) to detect if a string is a phone number.
In total, we sampled 70 tasks.

\paragraph{sampling programs  $P(f|T)$}
For each sampled task, we sample a number of \SP{} by prompting a LLM to generate code from the task, and we sample a number of \PP{} by prompting a LLM to generate a prompt that can output the answer directly.
Ideally we could also hire human programmers to write programs and the prompts for a more naturalistic set of programs.
The sampled tasks and the sampled programs together forms a corpus $\mathcal{M}$, and from this corpus $\mathcal{M}$ we can build the empirical performance prior $P(\theta_f)$.
For this work, We sampled 10 programs for either \SP{} and \PP{} per task, for a total of 700 \SP{} and 700 \PP{}.

\paragraph{sampling instances $P((x,y)|T)$ to estimate $\theta_f$}
Each task comes equipped with an instance sampler, $P((x,y)|T)$, which we can query for a large number of input $x$ and oracle labeled output $y$ (e.g. a large number of randomly generated strings, and whether the string is a valid phone number).
We approximate the true performance of a program $f$ by taking its average performance on the sampled instances.
\begin{equation}
\tilde{\theta}_f
\;:=\;
\frac{1}{N}\sum_{i=1}^{N}\mathbf{1}\!\big[f(x_i)=y_i\big],
\qquad (x_i,y_i)\sim p((x,y)\mid T).
\end{equation}
For this part, we sample $N = 100$ instances per task.
In total, we performed 70000 function calls to execute \SP{}, and 70000 LLM API-calls to execute \PP{}.

\paragraph{empirical performance prior for \SP{} and \PP{}}
The approximated performances of the 700 \SP{} and \PP{} are collated together to form the empirical performance prior for symbolic programs $P(\theta_{sp})$ and prompt programs $P(\theta_{pp})$ (Figure \ref{fig:sp-vs-pp-prior}).
The most striking difference is that \SP{} has a sharp, bi-modal distribution where a program is either all correct (1.0) or completely incorrect (0.0), whereas the prior for \PP{} is much more diffused, with a large number of ``near perfect'' programs.

\begin{figure}[t]
\centering

\begin{subfigure}[t]{0.48\linewidth}
\centering
\includegraphics[width=\linewidth]{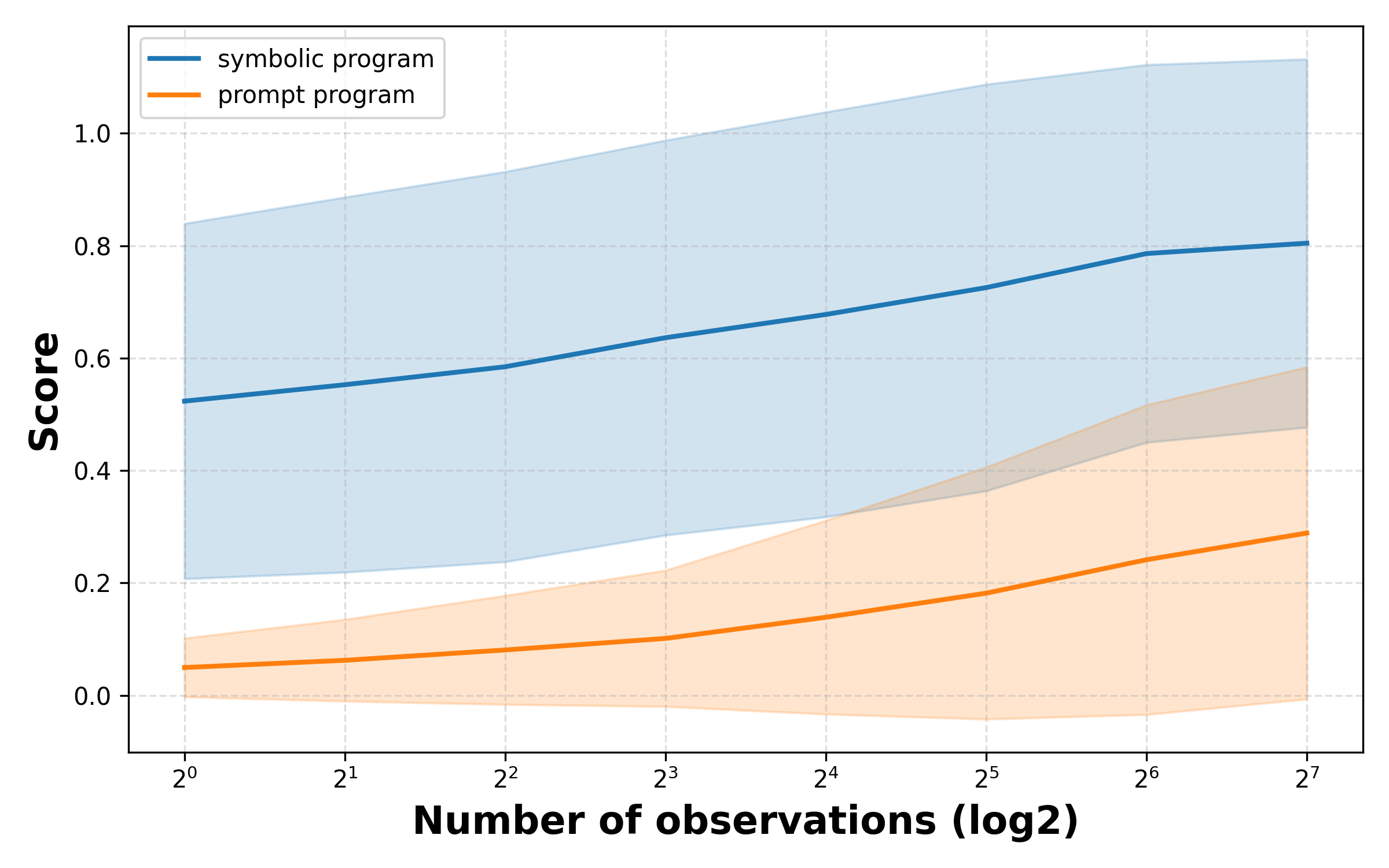}
\caption{Probability density of ground truth performance under posterior for symbolic $P(\theta_{sp}=\theta_{sp}^*\mid O)$ and prompt $P(\theta_{pp}=\theta_{pp}^*\mid O)$ programs. Predicting performance of symbolic programs is better.}
\label{fig:posterior-score-curve}
\end{subfigure}
\hfill
\begin{subfigure}[t]{0.48\linewidth}
\centering
\includegraphics[width=\linewidth]{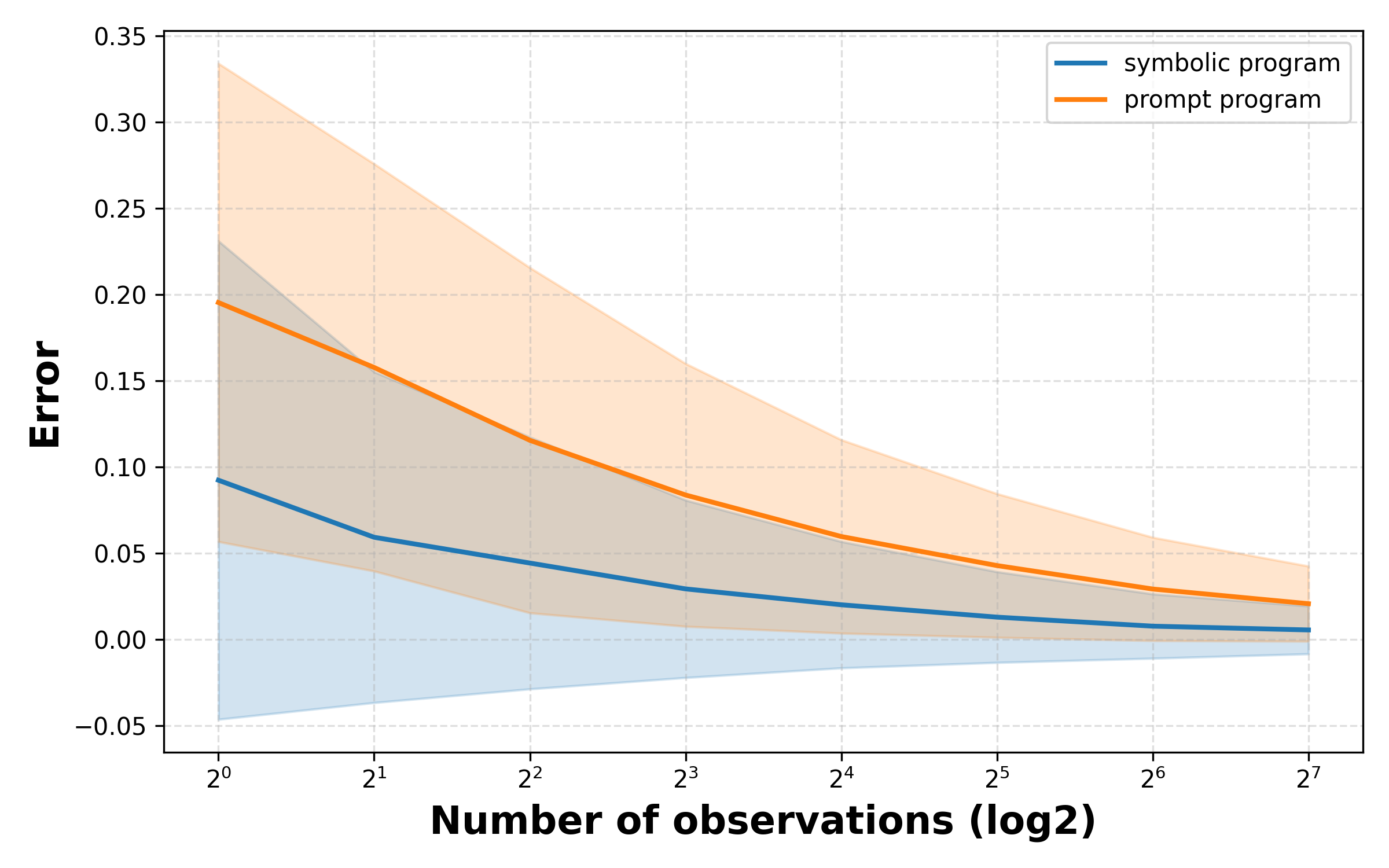}
\caption{Absolute difference error between predicted performance and ground-truth for symbolic $|\hat{\theta}_{sp}-\theta_{sp}^*|$ and prompt $|\hat{\theta}_{pp}-\theta_{pp}^*|$ programs. Predicting performance of symbolic programs achieves a lower error.}
\label{fig:posterior-error-curve}
\end{subfigure}

\caption{Evaluation of performance prediction as a function of observed examples. Over both metrics, it was easier to predict the performance for symbolic programs.}
\label{fig:posterior-score-error}
\end{figure}

\subsection{Using Prior for Performance Prediction}
From the empirical performance priors, we can answer our original question: ``Why observing a small number of successes guarantees good performance for symbolic programs but not prompt programs''.
In Figure \ref{fig:taifeng} we show the posterior performance distribution for both \SP{} and \PP{} after observing successes on 3 test cases.
As we can see, from the posterior for \SP{}, the mass is nearly concentrated on perfect performing programs, while the same cannot be said for \PP{} --- it has many ``near perfect'' programs simply getting lucky.

The above observation holds true in general. By sampling a program $f$ (either symbolic or prompt) from our empirical distribution, and trying to estimate its true performance from observations $P(\theta_f|O)$, we get the following results in~\Cref{fig:posterior-score-error}.
As we can see, it is much more difficult to predict the true performance for \PP{} compared to \SP{}

\paragraph{conclusion} We conclude that the difference in performance prediction can be largely attributed to the distinct shapes of the \SP{} and \PP{} performance priors. 
\textit{The prior for symbolic programs affords us a confident prediction from few examples, while the prior for prompt programs does not}.


\section{RAP: Retrieved Approximate Prior}
\label{sec:RAP}

\begin{figure}[t!]
    \centering
    \includegraphics[width=0.8\columnwidth]{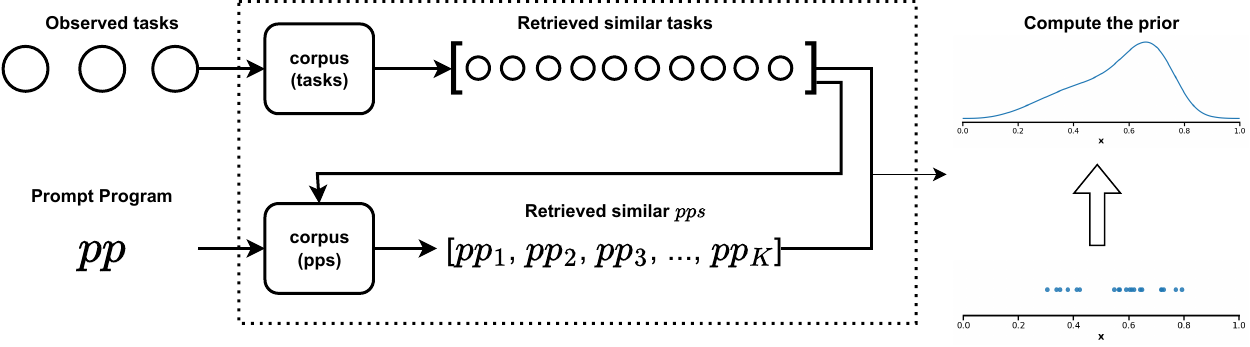}
    \caption{
        The overview of constructing prior. RAP retrieves tasks similar to the observed examples and prompt programs similar to the target prompt program. The retrieved programs are evaluated on the retrieved tasks to obtain empirical success rates, shown as blue points in the lower-right panel. These empirical performances are then smoothed into a mixture of Beta distributions, which serves as the task-specific performance prior.
    }
    \label{fig:rap-overview}
\end{figure}

To construct a more informative empirical prior for performance prediction, we introduce \emph{Retrieved Approximate Prior (RAP)}. RAP takes as input (1) a target prompt program $\ppt$ (2) a set of observed examples from the target domain $\mathcal{D} = \{x_1, \dots x_k\}$, and (3) an \emph{arbitrary} corpus $\mathcal{M}=(\mathcal{M}_{tasks}, \mathcal{M}_{pps})$ of tasks and prompt programs. 
RAP first constructs an informative prior $P(\theta_{\ppt} \mid \ppt, \mathcal{D}, \mathcal{M})$ by retrieving related tasks and prompt programs from the corpus (\Cref{fig:rap-overview}). 
RAP then executes $\ppt$ on $\mathcal{D}$ to get the pass/fail outcome $O$.
Finally, RAP performs Bayesian update on the retrieved prior with $O$ to produce the posterior that predicts performance:

\begin{equation}
P_{\text{RAP}}(\theta_{pp_{\text{target}}} \mid \ppt, \mathcal{D}, \mathcal{M})
\propto
P(O \mid \theta_{pp_{\text{target}}})
P(\theta_{pp_{\text{target}}} \mid \ppt, \mathcal{D}, \mathcal{M}).
\label{eq:rap-posterior}
\end{equation}

\subsection{Retrieve Similar Tasks}
Given an observed task set $\mathcal{D}$, we retrieve an extended task set $\mathcal{D}'$.
We do so by retrieving the top-n most similar tasks in $\mathcal{M}_{tasks}$ for each $x' \in \mathcal{D}$, where similarity is measured using a textual embedding function $e(\cdot)$. These retrieved tasks serve as a proxy for target domain.
\begin{equation}
\mathcal{D}'
=
\bigcup_{x' \in \mathcal{D}}
\operatorname{Top\text{-}n}\Big(
\{x \in \mathcal{M}_{tasks}\},
\; \cos\!\big(e(x), e(x')\big)
\Big),
\label{eq:task-retrieval}
\end{equation}

\subsection{Retrieve Similar Prompt Programs}
RAP retrieves a top-K set of prompt programs that are similar to the $\ppt$ in terms of \emph{behavior} --- the number of tasks in $\mathcal{D}'$ where both programs agree (both correct or incorrect), $s_{\mathcal{D}'}(\ppt,pp_i)$.

\begin{equation}
\label{eq:pp-retrieval-and-prior}
\begin{aligned}
\mathcal{PP}\prime
&:= \operatorname{Top\text{-}K}\Big(\{pp_i \in \mathcal{M}_{pps}\},\; s_{\mathcal{D}'}(\ppt,pp_i)\Big),\\
\end{aligned}
\end{equation}

\subsection{Constructing Prior and Posterior Update}

After retrieving similar tasks $\mathcal{D}'$ and prompt programs $\mathcal{PP}'$,
we construct the distribution of the unknown success rate $\theta_{\ppt}$ as a mixture of Beta distribution.

Let $a_i$ and $b_i$ denote the numbers of successes and failures of $pp_i$ on $\mathcal{D}'$.
We set $\alpha_i=a_i+1$ and
$\beta_i=b_i+1$. 
Thus,
$pp_i$ induces the component $\mathrm{Beta}(\theta \mid \alpha_i,\beta_i).$ Aggregating the evidence from the top-$K$ retrieved programs yields the
retrieved prior as a mixture of these components: 
\begin{equation}
P(\theta)
=
\sum_{i=1}^{K}
\frac{1}{K} \,
\mathrm{Beta}(\theta \mid \alpha_i, \beta_i).
\end{equation}

Given observed outcomes $O$, let $a$ and $b$ denote the numbers of successes
and failures. Since each mixture component is a Beta prior over the same
Bernoulli success rate $\theta$, the observed outcomes update each component
by conjugacy to produce the updated posterior distribution:

\begin{equation}
P(\theta \mid O,\mathcal{M})
=
\sum_{i=1}^{K}
\frac{1}{K}
\,
\mathrm{Beta}(\theta \mid \alpha_i+a,\beta_i+b),
\label{eq:rap-updated-mixture}
\end{equation}

\paragraph{Component-wise Prior Strength Adjustment.}

It is possible that the top-K retrieved $pp_i$ behave drastically differently than $\ppt$ on the target domain.
To resolve this, RAP adjust the strength of each Beta component's parameters based on performance similarity on in-domain tasks.

Let $P_i(\theta)=\mathrm{Beta}(\theta \mid \alpha_i,\beta_i)$ denote the prior component induced by retrieved program $pp_i$. Let $P_{\text{obs}}(\theta) = \mathrm{Beta}(\theta \mid a+1,b+1)$ denote the Beta distribution induced directly from in-domain examples without the corpus. We compute a positive scaling factor based on a normalized Earth Mover's Distance (EMD):
\begin{equation}
\lambda_i
=
1 -
\mathrm{EMD}\!\left(
P_i,
P_{\text{obs}}
\right),
\quad \lambda_i \in [0,1].
\label{eq:component-similarity}
\end{equation}

We then multiply this scaling factor $\lambda_i$ to the Beta distribution's parameters.
In addition, we also cap the concentration value to ensure $\lambda_i\alpha_i + \lambda_i\beta_i < c_{\max}$, where we set $c_{\max} = 40$.
Finally:

\begin{equation}
P_{RAP}(\theta \mid O)
=
\sum_{i=1}^{K}
\frac{1}{K} \,
\mathrm{Beta}(\theta \mid \frac{\alpha_i}{C_i} + a, \frac{\beta_i}{C_i} + b),
~~~ \text{Where } C_i = \frac{\alpha_i + \beta_i}{\min(\lambda_i (\alpha_i + \beta_i), c_{max})}.
\end{equation}


Note that as the number of in-domain observations increases, the likelihood term --- $a$ and $b$ --- dominates the posterior update, ensuring convergence toward the empirical posterior induced purely by observed in-domain data.


\begin{figure}[t!]
    \centering
    \includegraphics[width=1.0\linewidth]{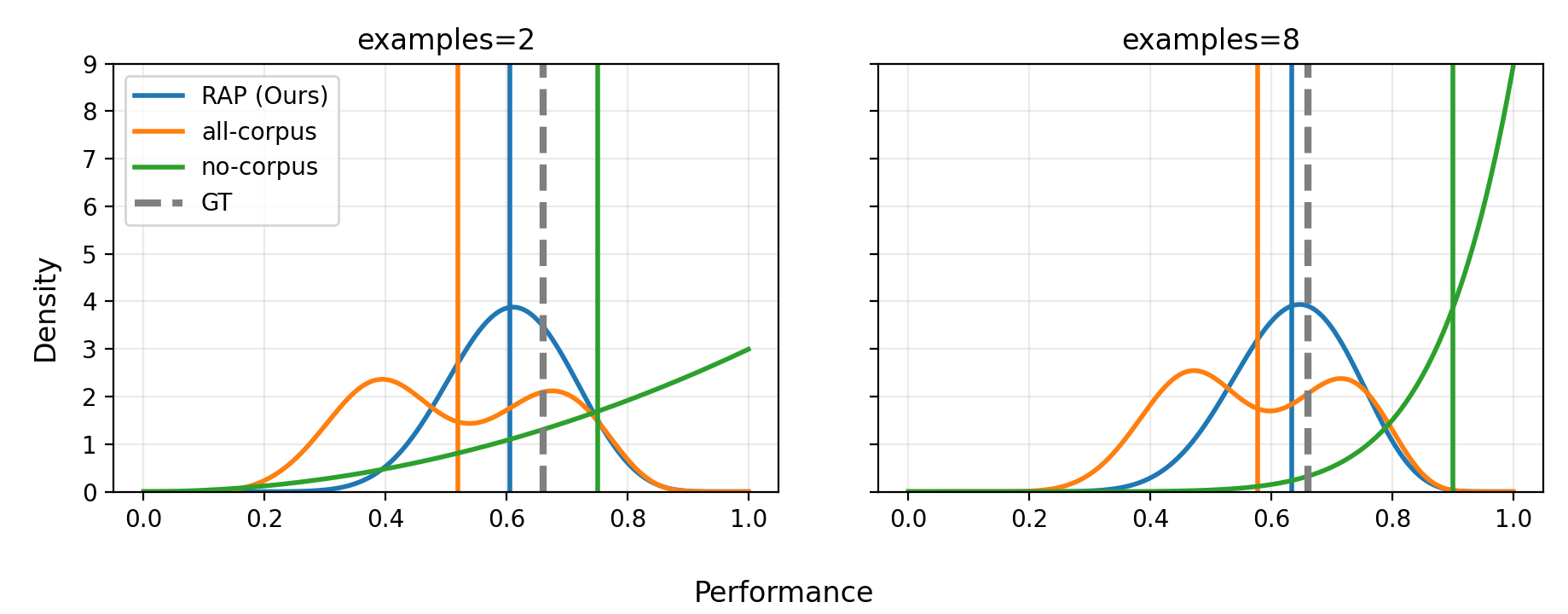}
    \caption{Our method constructs a more informative and adaptable prior than the baselines. Compared to \emph{all-corpus}, whose initial prior is poorly calibrated for the target domain, and \emph{no-corpus}, where the first few observations can excessively distort the posterior, our approach provides a better-calibrated starting point that can be effectively updated as new observations are incorporated. The vertical line marks the posterior mean used as the performance prediction; closer alignment with the ground-truth performance indicates better calibration.}
    \label{fig:qualitative}
\end{figure}

\setlength{\abovecaptionskip}{2pt}
\setlength{\belowcaptionskip}{2pt}

\begin{figure*}[!t]
  \centering

  \begin{subfigure}{0.48\textwidth}
    \centering
    \includegraphics[width=0.89\linewidth]{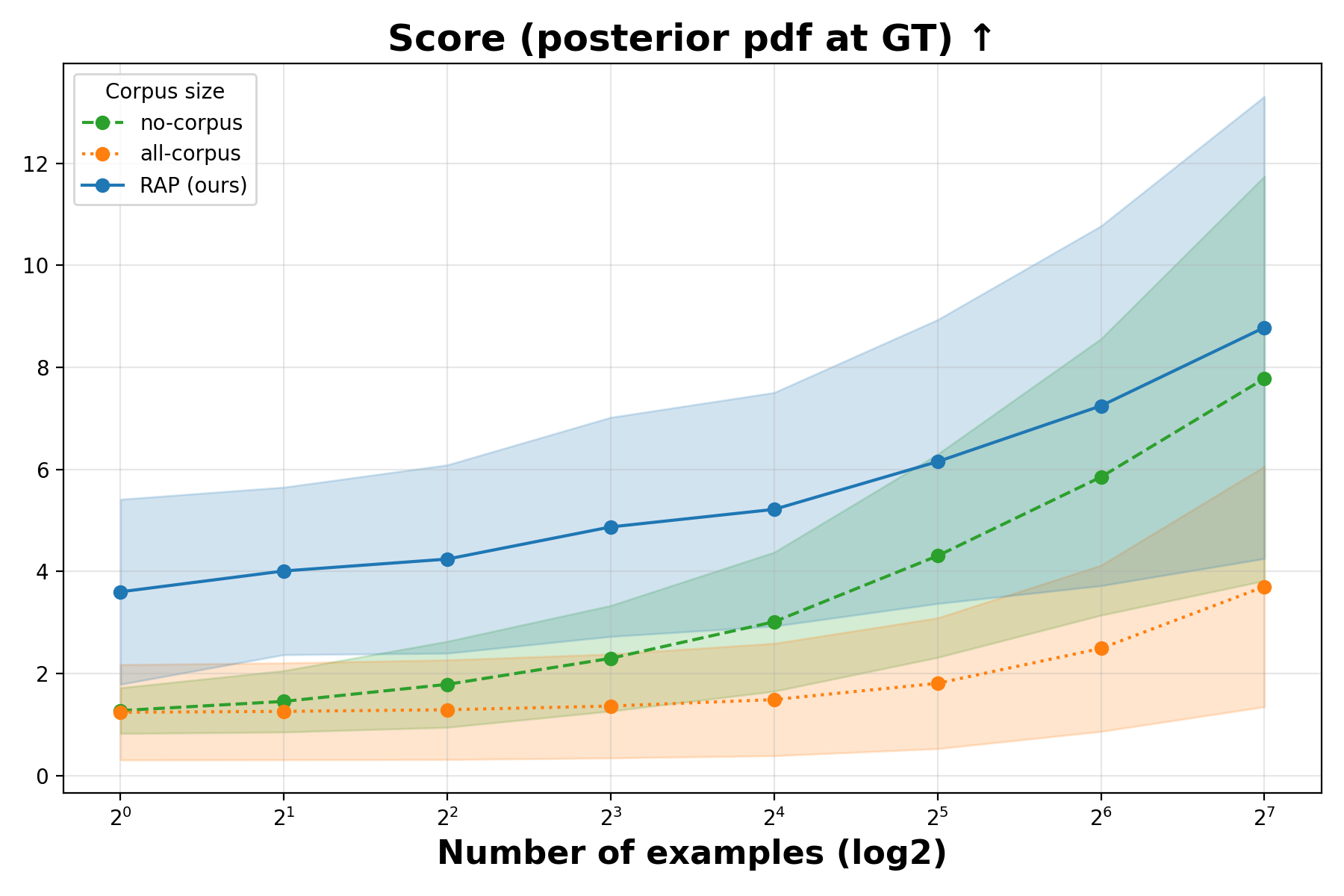}
    \includegraphics[width=0.89\linewidth]{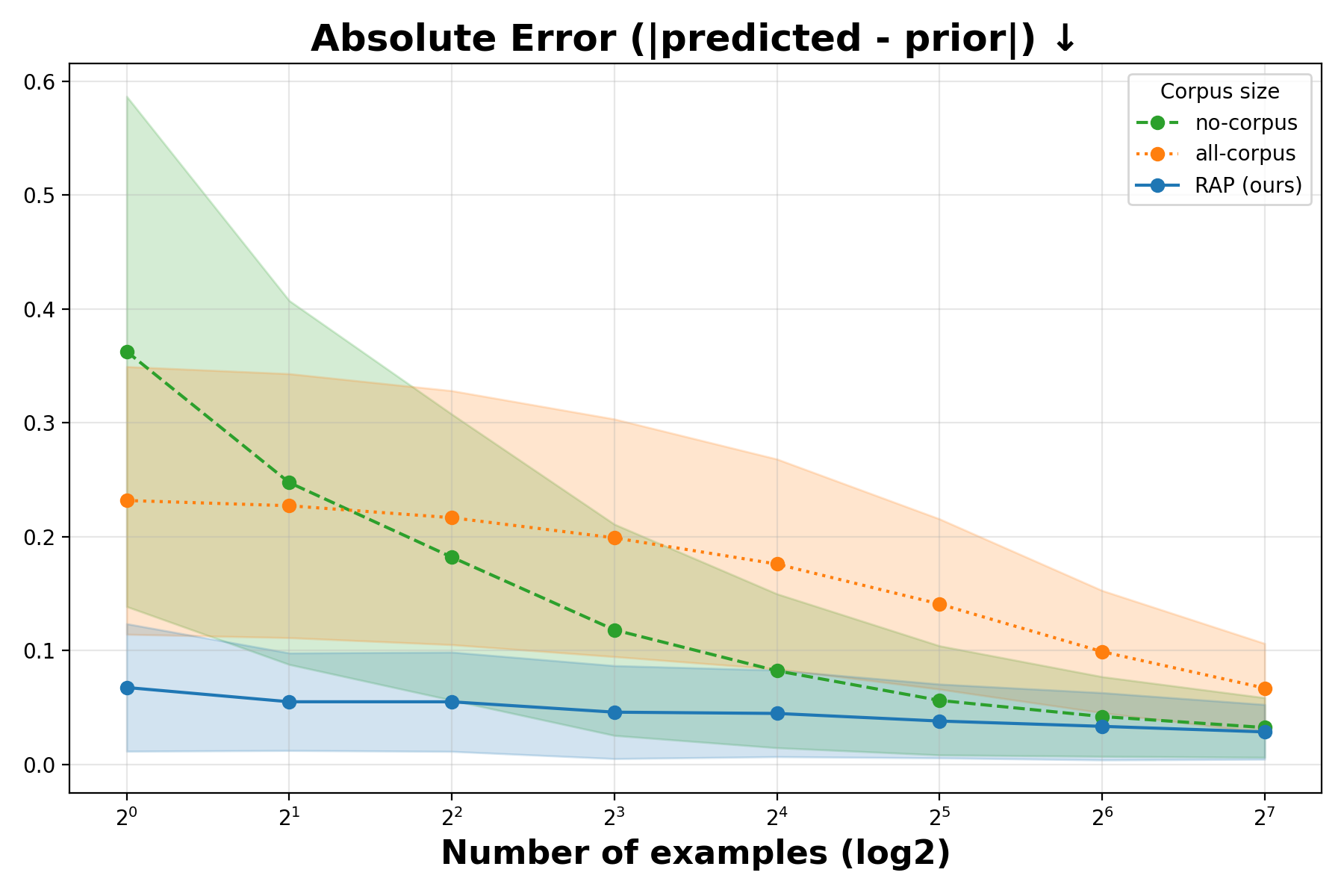}
    \caption{in-domain}
    \label{fig:score-in}
  \end{subfigure}\hfill
  \begin{subfigure}{0.48\textwidth}
    \centering
    \includegraphics[width=0.89\linewidth]{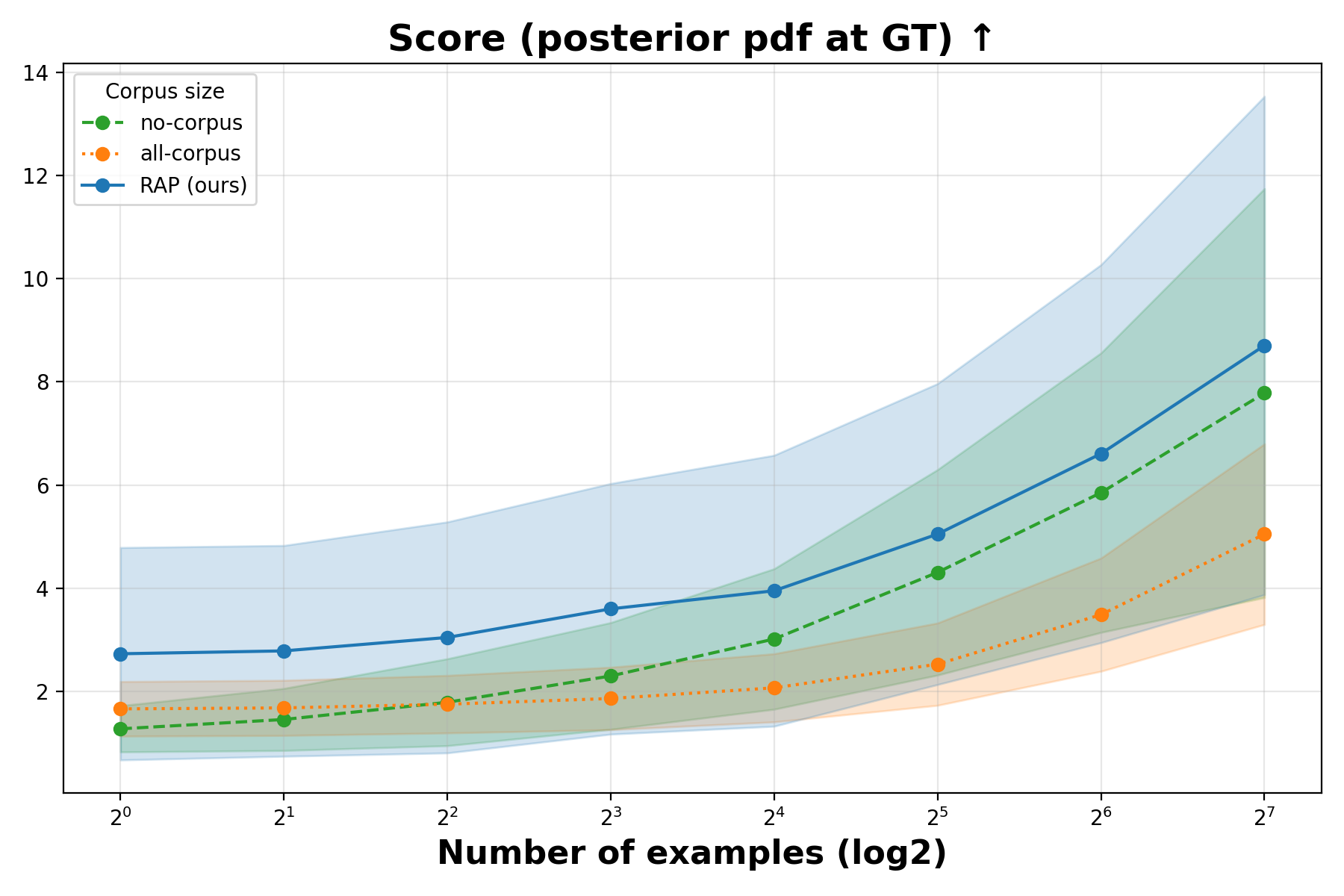}
    \includegraphics[width=0.89\linewidth]{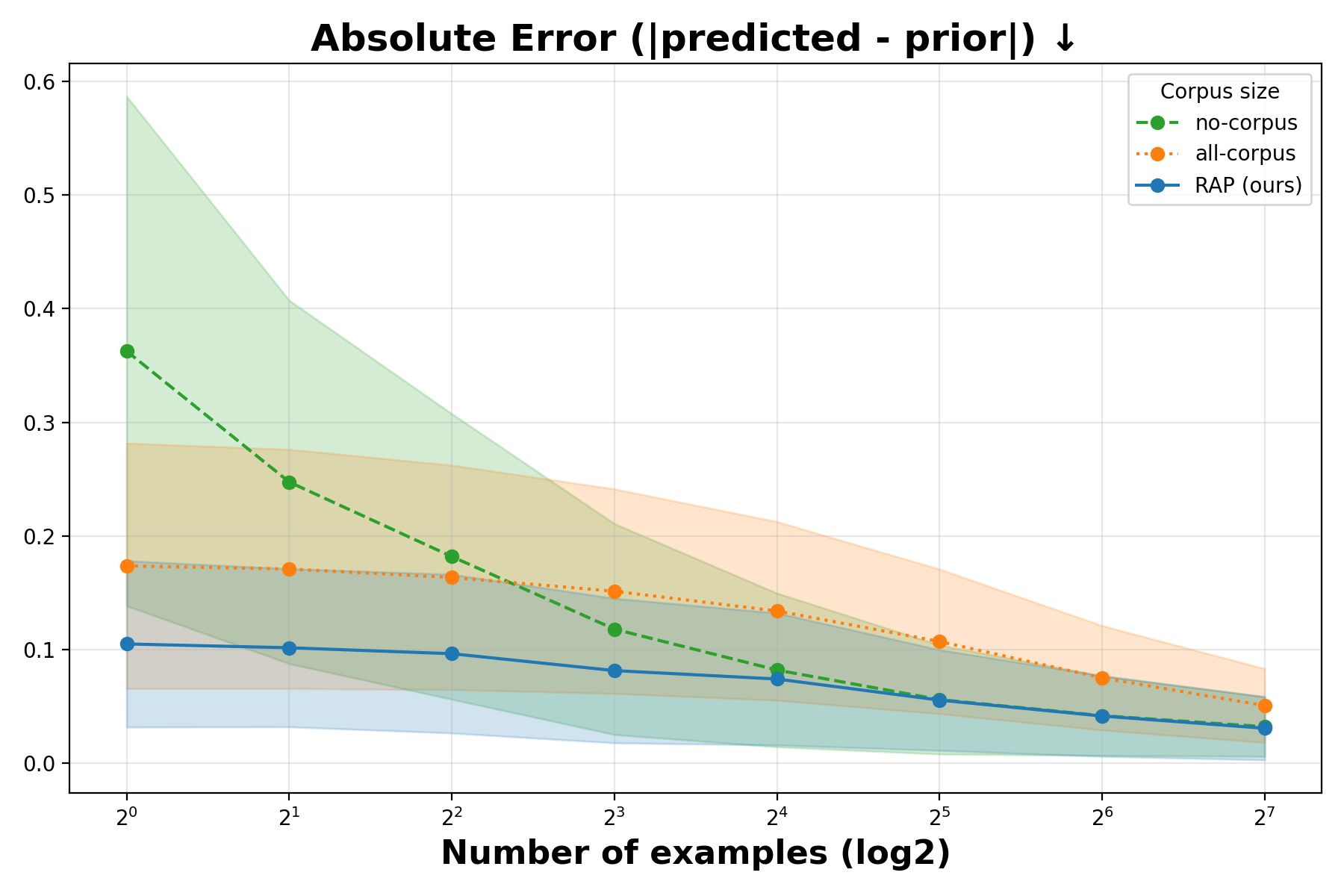}
    \caption{out-of-domain}
    \label{fig:score-out}
  \end{subfigure}\hfill

  \caption{RAP out performs baselines in both in-domain and out-of-domain settings. In the in-domain setting, the corpus contains tasks from the target domain, enabling RAP to construct a more informative prior, which leads to higher performances.}
  \label{fig:muscle-2x2}
\end{figure*}

\section{Experiment}
\label{sec:experiment}
We validate the following:
\begin{compactenum}
    \item RAP outperforms baselines that either do not use a corpus (no-corpus) or naively use the entire corpus (all-corpus), while effectively leveraging in-domain tasks and remaining robust to irrelevant out-of-domain tasks in the corpus (\cref{fig:muscle-2x2}).
    \item Performance of RAP increases as the sizes of corpus increases (\cref{fig:posterior_score_curves_big})
    \item RAP converges gracefully to the in-domain posterior when given a number of in-domain examples (\cref{fig:qualitative})
    \item RAP is robust to hyperparameter choices (Appendix \cref{app:hyperparameter-ablation}).
\end{compactenum}

\subsection{Methodology}

\paragraph{Benchmarks}

We chose benchmarks from domains in Appendix \cref{app:benchmarks} \Cref{tab:domains_for_pp}, covering the range of tasks that people generally use prompt program for --- knowledge and reasoning based multiple choice, general Python code generation, and embodied agent.
These domains are \emph{different} than the domains considered in Section 3 in the following way --- rather writing a different program per task, the goal here is to write a general \PP{} that can solve many different tasks.
For instance, write a single prompt that can generate multiple Python solutions for the MBPP benchmark.
No symbolic program will work for the kind of tasks considered in this section.

\paragraph{Methods}
In this paper we consider three methods.

\begin{compactitem}
    \item \textbf{no-corpus}: Uses a fixed $\mathrm{Beta}(1,1)$ prior, which is uniform and independent of the corpus.
    \item \textbf{all-corpus}: Uses the performance prior estimated from the entire corpus.
    \item \textbf{RAP (ours)}: As described in Section \ref{sec:RAP}.
\end{compactitem}

\paragraph{Evaluation}

The prediction algorithm is presented with a prompt program \PP{} and few in-domain tasks $T_1, \dots, T_K$ from a \emph{target domain}, and gives a posterior distribution on the performance of \PP{} on future tasks from the target domain.

We report two metrics: the probability density of the posterior at the ground-truth performance (\textbf{SC}), and the absolute error (\textbf{AE}). They are described in ~\Cref{sec:predict_performance}.
We consider two settings, in-domain and out-of-domain. In the in-domain setting, the corpus $\mathcal{M}$ contain tasks from the target domain.
For instance, if we wish to predict performance of a \PP{} in the \texttt{math} domain, we will put some other tasks from the \texttt{math} domain in the corpus.
In the out-of-domain setting, no tasks from the same domain are included in the corpus.
For both settings we perform leave-one-out evaluation, where we single out each domain (e.g. \texttt{math}, \texttt{MBPP}, ...) as the target domain.

We evaluate 60 prompt programs under different settings, varying the language model, prompt, and execution mechanism \footnote{Note that we consider a prompt program \emph{inclusive} of its prompt, its LLM, its execution mechanisms, etc}. We use performance measured on 200 held-out tasks to approximate the ground-truth performance. Additional details are provided in the appendix.

\subsection{Results}

\paragraph{Qualitative Result}
We first qualitatively compare our algorithm with the baselines in~\Cref{fig:qualitative}. Our method produces a more informative yet flexible prior, which leads to more reasonable posterior updates than the baselines. Our algorithm also approaches the no-prior baseline, as the likelihood term dominates for a large number of examples.

\paragraph{Quantitative Result}
Results in~\Cref{fig:muscle-2x2} compare the performance in in-domain and out-of-domain settings. RAP performs best in both settings. Further, using in-domain examples consistently yields better performance, especially when only a limited number of examples are available.

\paragraph{Ablation Study} Appendix~\ref{app:additional-results} shows that RAP remains robust across retrieval hyperparameters, prior concentration choices, corpus sizes, and different LLM backbones.

\section{Related Works}


\textbf{Reliability of symbolic and prompt programs.} Symbolic programs are amenable to \emph{mechanical} correctness checks. Reliability can be increased via a simple propose-and-reject loop—generate multiple \SP{} candidates and accept only those that satisfy the checker~\citep{alur2013syntax, solar2008program}. Prompt programs are difficult to certify because “execution” is stochastic and implicit--correctness typically cannot be decided by a deterministic verifier. As a result, reliability is often improved with sampling-and-selection heuristics (e.g., self-consistency~\citep{wang2022self} / best-of-$N$~\citep{wang2024math, ouyang2022training}) and with empirical robustness evaluation~\citep{srivastava2023beyond, liang2022holistic}.

\textbf{Performance prediction.}
One line of work studies \emph{sensitivity}: small paraphrastic or formatting changes can cause substantial variance of performance ~\citep{razavi2025benchmarking}. A second line focuses on \emph{prompt optimization}, improving accuracy through more sample-efficient optimization and selection strategies~\citep{agrawal2025gepa, shi2024efficient}. 
Broader \emph{performance prediction} line of works aim to rank relative strengths of different LLMs using a smaller benchmark ~\citep{wang2025rethinking, ye2023predictable}, and to anticipate when an LLM may fail ~\citep{pacchiardi2025predictaboard}.
Our work is the first to take a simple coin-toss model, and clearly quantifies the difference between \SP{} and \PP{} using this model.
\section{Conclusion, Limitations, and Future Works}
\label{limitation}
We study program performance prediction by: 1) Framing prediction as posterior inference $P(\theta_f \mid O)$, 2) Show that the empirical priors differ sharply between symbolic (\SP{}) and prompt (\PP{}) programs, and 3) Introduce RAP, which predicts \PP{} performance from few examples by approximating an in-domain prior using an arbitrary corpus.
We consider RAP a first line of attack on the problem of performance prediction, and it is far from perfect.
Chiefly, the top-K retrievals of relevant tasks and \PP{}s would be expensive as the sizes of the corpus grows.
In future work, we may explore actively querying the right (task, \PP{}) pair to execute, and by building a low-ranking approximation of the corpus of $\mathcal{M} = tasks \times pps$.

\bibliography{example_paper}
\bibliographystyle{icml2026}
\clearpage
\appendix
\section{Experiments Details}

\subsection{Hyperparameters Setting}
RAP involves three hyperparameters in our experiments: the number of retrieved tasks, set to $n=100$; the number of retrieved prompt programs, set to $K=5$; and the maximum prior concentration, set to $c_{\max}=40$.

\subsection{Computational Cost}
Constructing the full prior from scratch for all experiments costs approximately \$300 in API usage, including the evaluation of prompt programs over the task corpus and the observed example sets.

\subsection{The formula for All-Corpus}
For the all-corpus baseline, every prompt program $pp_m \in \mathcal{P}$ is evaluated on all corpus tasks. This yields $\alpha_m$ successes and $\beta_m$ failures for each prompt program. Given the target prompt program's observed outcomes on the in-domain examples, let $s$ and $f$ denote its numbers of successes and failures, respectively. The all-corpus posterior is then defined as:
\begin{equation}
p_{\text{all-corpus}}(\theta)
=
\frac{1}{|\mathcal{P}|}
\sum_{pp_m \in \mathcal{P}}
\mathrm{Beta}\left(
\theta \mid \alpha_m + s,\ \beta_m + f
\right).
\end{equation}

We also refer to this baseline as \emph{all-prior}, since it uses all prompt programs in the corpus to construct the prior before incorporating the target program's observed outcomes.

\subsection{Benchmark Details}
\label{app:benchmarks}
Our benchmarks cover three broad classes of prompt-programming tasks: knowledge-intensive question answering, code generation, and embodied agentic tasks. Specifically, we use six MMLU-PRO domains, two code-generation benchmarks, and two ALFWorld task types. These benchmarks provide a diverse testbed for evaluating whether RAP can construct useful task-specific priors across different forms of prompt program execution.

\begin{table}[h!]
\centering
\small
\setlength{\tabcolsep}{6pt}
\renewcommand{\arraystretch}{1.12}
\caption{Benchmark domains used in our experiments. We may separate domains from MMLU\_PRO and Alfworld. For most domains, three datasets are provided: a test set (200 tasks), an example set (128 tasks), and a question pool (400 tasks). More details can be found in appendix.}
\label{tab:domains_for_pp}
\begin{tabular}{@{}l l@{} l}
\toprule
\textbf{Type} & \textbf{Domain}  & \textbf{\#Tasks}\\
\midrule
\multirow{6}{*}{\text{Knowledge (MMLU\_PRO)}}
& \texttt{math} & \texttt{728}\\
& \texttt{physics} & \texttt{728}\\
& \texttt{chemistry}  & \texttt{728}\\
& \texttt{economics}  & \texttt{728}\\
& \texttt{law}  & \texttt{728}\\
& \texttt{engineering}  & \texttt{728}\\
\addlinespace
\multirow{2}{*}{Code Generation}
& \texttt{MBPP}  & \texttt{728}\\
& \texttt{HumanEval}  & \texttt{168}\\
\addlinespace
\multirow{2}{*}{Agentic (Alfworld)}
& \texttt{pick and place}  & \texttt{448}\\
& \texttt{cool then place}  & \texttt{448}\\
\bottomrule
\end{tabular}

\end{table}

\subsection{Part 1: Comparison Between Symbolic Programs and Prompt Programs}
In this part, we investigate a diverse set of tasks spanning multiple domains. Some tasks do not naturally provide more than 100 instances; for these cases, we augment or complete the task instances to ensure a sufficiently large evaluation set (approximately 100+ instances) so that performance estimates are statistically reliable.

For inference on tasks specified by few-shot examples, we use GPT-5-mini as the underlying language model to ensure strong and stable prompt program performance. For all other settings, we use GPT-4o-mini or other commonly adopted language model configurations. Table~\ref{tab:sp_pp_tasks} summarizes the task specifications and model choices used in this part.

\begin{table}[h!]
\centering
\caption{Benchmark tasks used in Part~1, including task specification format and transduction models.}
\label{tab:sp_pp_tasks}

\setlength{\tabcolsep}{3pt}   
\small

\begin{tabular}{@{} l l l c @{}}
\toprule
\textbf{Benchmark} & \textbf{Task Specification} & \textbf{Transduction Model} & \textbf{Data Augmented} \\
\midrule
Structured Regex & Natural language & GPT-4o-mini & Yes \\
HumanEval & Natural language & GPT-4o-mini & Yes \\
ARC & Few-shot examples & Llama-3.1 (fine-tune) & No \\
List & Few-shot examples & GPT-5-mini & Yes \\
\bottomrule
\end{tabular}
\end{table}

\subsection{Part 2: The details of pp only experiments}
In the second part of the experiments, we evaluate 60 prompt programs, including 24 from multiple-choice question answering, 24 from code generation, and 12 from embodied AI tasks. We consider two language models, GPT-4o-mini and GPT-5-mini, and evaluate two prompting strategies: chain-of-thought (CoT), which explicitly elicits intermediate reasoning before producing an answer, and a vanilla setting, which does not allow the model to output reasoning content.

For each domain, we use the official checker to verify correctness. Most reported success rates are consistent with the official records.

Task construction proceeds as follows. The test set is used to obtain ground-truth performance, the example set is sampled to represent observed tasks, and the question pool is used to construct the task corpus. For the HumanEval domain, due to the limited number of tasks, the dataset is used exclusively as the question pool. For agentic domains, the size of the question pool is reduced because executing multi-turn environments is computationally expensive.

We construct prompt programs using commonly adopted prompt-design paradigms for each benchmark category. In particular, we instantiate task-specific instruction templates for multiple-choice question answering, code generation, and embodied agentic tasks, while ignoring fixed formatting details imposed by the prompt-programming Dspy framework. Representative examples are shown below.

\paragraph{MMLU-Pro.}
For multiple-choice question answering, the prompt encourages careful reasoning over the question and answer options:
\begin{tcolorbox}[colback=gray!5, colframe=gray!40, boxrule=0.5pt, arc=2pt, left=4pt, right=4pt, top=4pt, bottom=4pt]
Analyze the multiple-choice question carefully. Reason step by step, identify the key concept, evaluate each option, eliminate weaker choices, and select the single best answer. Return only the final answer in the required format.
\end{tcolorbox}

\paragraph{MBPP.}
For code-generation tasks, the prompt instructs the model to produce a correct and readable Python function:
\begin{tcolorbox}[colback=gray!5, colframe=gray!40, boxrule=0.5pt, arc=2pt, left=4pt, right=4pt, top=4pt, bottom=4pt]
Write a clean, efficient Python function that solves the problem and passes all tests. Read the description, constraints, examples, and assertions carefully, handle edge cases, and ensure correctness and readability.
\end{tcolorbox}

\paragraph{ALFWorld.}
For embodied agentic tasks, the prompt specifies the task context and asks the model to select the next action:
\begin{tcolorbox}[colback=gray!5, colframe=gray!40, boxrule=0.5pt, arc=2pt, left=4pt, right=4pt, top=4pt, bottom=4pt]
You are an embodied household agent in an interactive environment. Your goal is to complete the given task description by selecting the best next action based on the task description, current observation, previous trajectory, and possible actions.
\end{tcolorbox}

\clearpage
\section{Ablation Study}
\label{app:additional-results}

\subsection{Hyperparameter Ablation Study}
\label{app:hyperparameter-ablation}
\paragraph{Retrieval Hyperparameters.}
We ablate the two retrieval hyperparameters used by RAP: the number of retrieved prompt programs $K$ and the number of retrieved tasks $n$. Increasing $K$ can provide more diverse prior components, but may also introduce irrelevant priors when the retrieved programs are not well aligned with the target program. In contrast, increasing $n$ generally improves performance by providing more task-level evidence for constructing a stable retrieval context.

\begin{figure}[h]
\centering
\begin{subfigure}{0.48\linewidth}
    \centering
    \includegraphics[width=\linewidth]{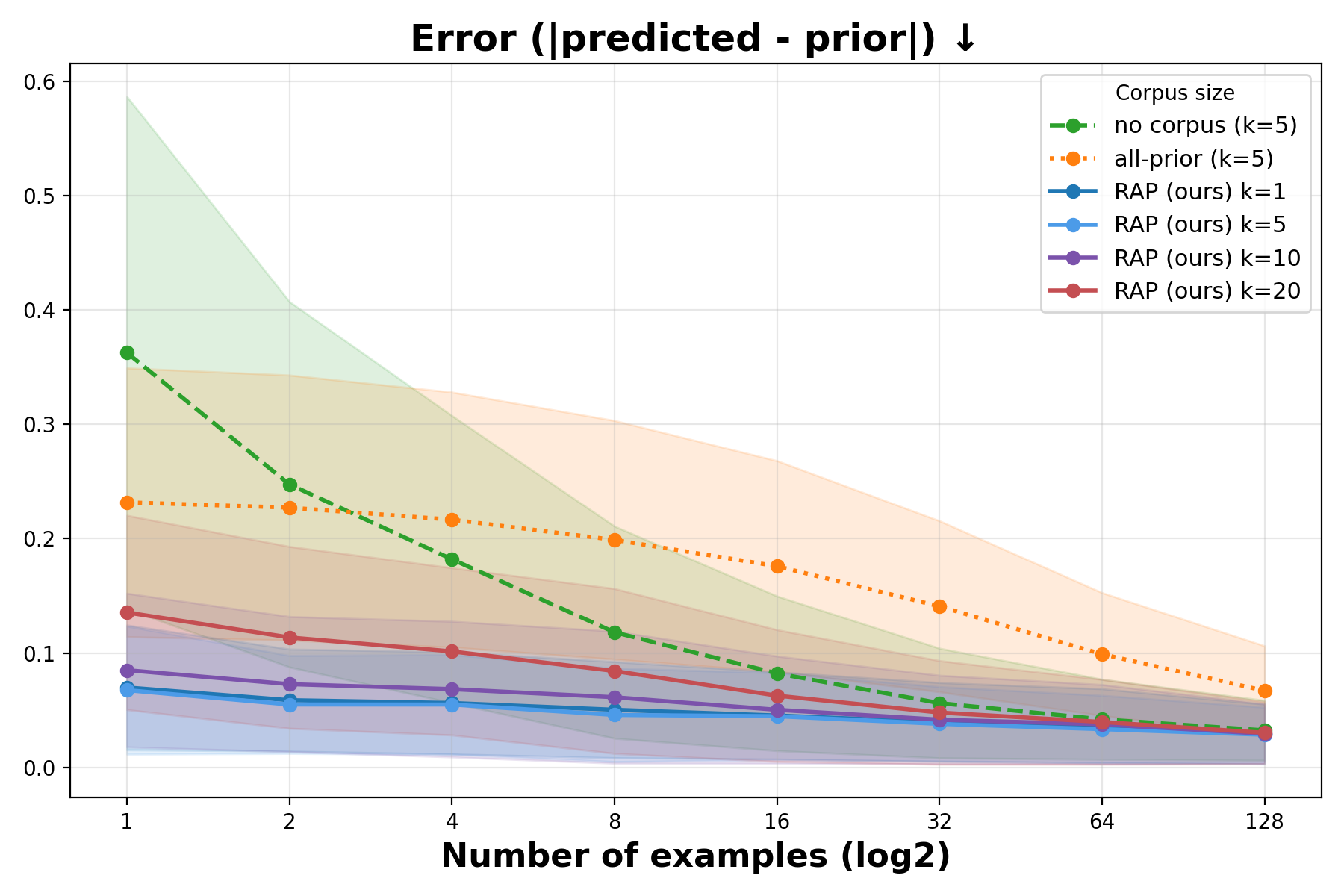}
    \caption{Number of retrieved prompt programs $K$.}
    \label{fig:ablation-k}
\end{subfigure}
\hfill
\begin{subfigure}{0.48\linewidth}
    \centering
    \includegraphics[width=\linewidth]{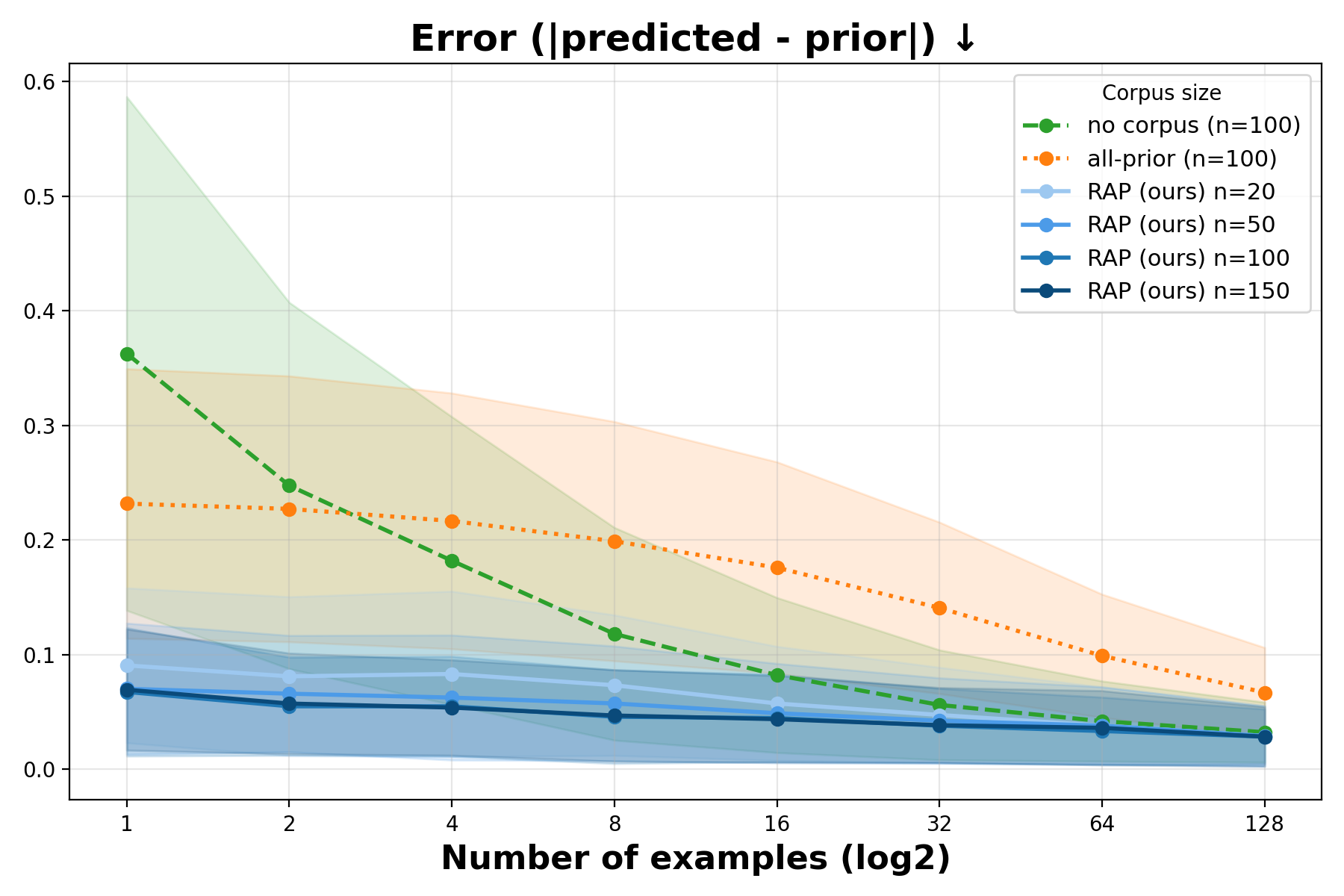}
    \caption{Number of retrieved tasks $n$.}
    \label{fig:ablation-n}
\end{subfigure}
\caption{Hyperparameter ablation for RAP. Left: increasing $K$ can improve prior diversity but may also introduce irrelevant prior components. Right: increasing $n$ generally improves prediction accuracy by stabilizing the retrieval context.}
\label{fig:hyperparameter-ablation}
\end{figure}

\paragraph{Effect of Prior Concentration Cap.}
We also ablate the maximum prior concentration $c_{\max}$, which controls the maximum strength assigned to each retrieved Beta prior component after similarity-based rescaling. A small $c_{\max}$ makes the retrieved prior easier to update with observed examples, while a large $c_{\max}$ allows the retrieved prior to have stronger influence but may make the posterior overconfident when retrieval is imperfect.

\begin{figure}[h]
\centering
\includegraphics[width=0.55\linewidth]{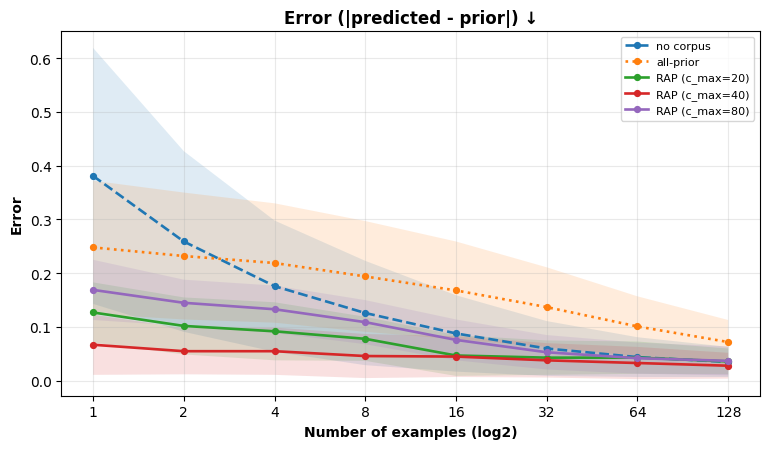}
\caption{Ablation over the maximum prior concentration $c_{\max}$. Moderate values provide a good balance between using retrieved prior information and remaining adaptable to observed in-domain examples.}
\label{fig:ablation-cmax}
\end{figure}


\subsection{Effects of Corpus Size}
\begin{figure}[h!]
\centering

\begin{subfigure}[t]{0.48\textwidth}
  \centering
  \includegraphics[width=1.0\linewidth]{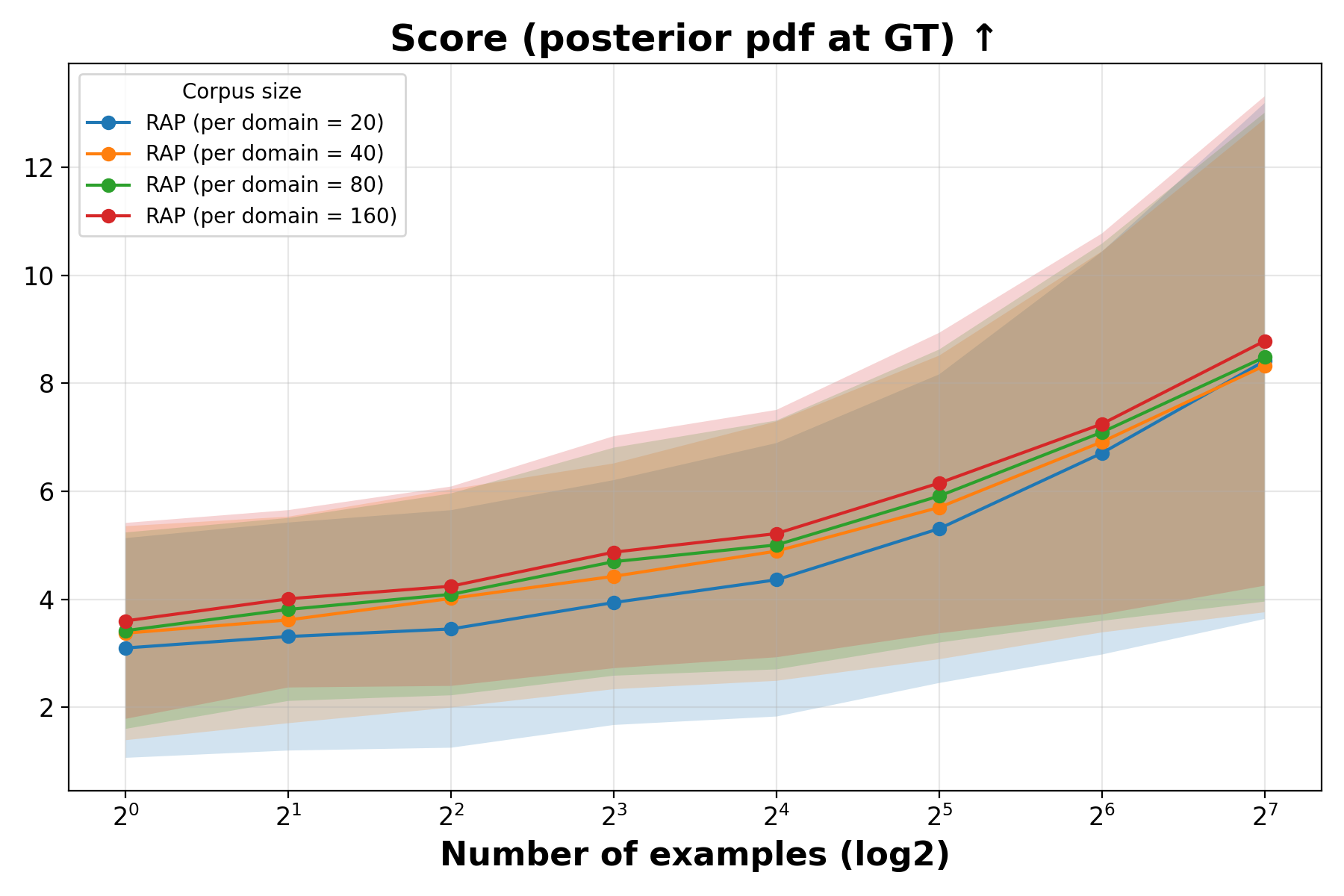}
  \caption{Posterior score curve (in-domain)}
  \label{fig:posterior_score_curve_in_domain}
\end{subfigure}\hfill
\begin{subfigure}[t]{0.48\textwidth}
  \centering
  \includegraphics[width=1.0\linewidth]{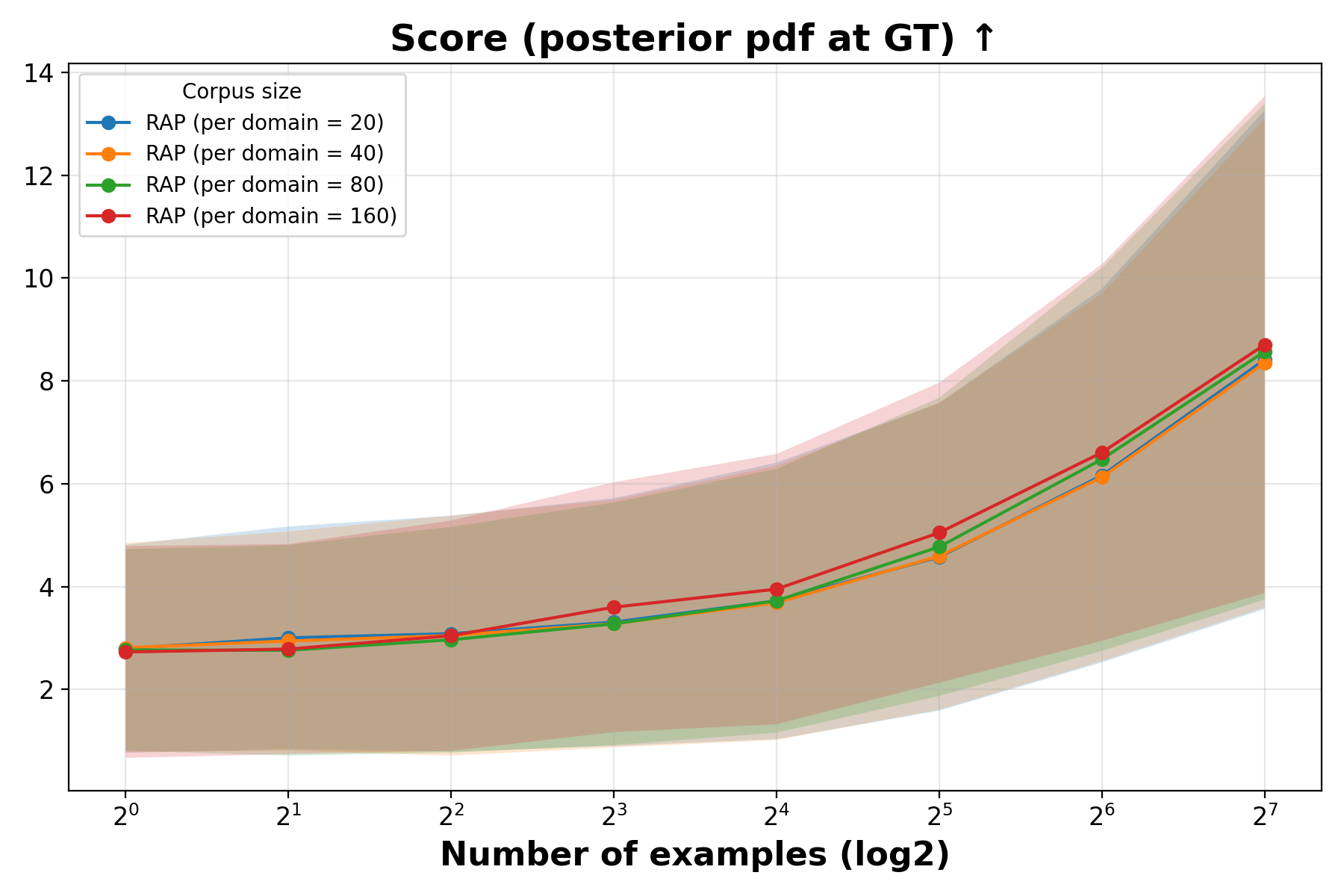}
  \caption{Posterior score curve (out-of-domain)}
  \label{fig:posterior_score_curve_out_domain}
\end{subfigure}

\caption{RAP benefits from a larger corpus size, where a larger corpus boosts performance in the in-domain setting (a) and does not harm performance in out-of-domain setting (b)}
\label{fig:posterior_score_curves_big}
\end{figure}

Intuitively, the larger the corpus size, the more likely our algorithm will be able to retrieve similar tasks and \PP{}s. 
We generate a sequence of corpora with increasing sizes and evaluate RAP's performance. 
Results are shown in~\Cref{fig:posterior_score_curves_big}. 
In both the in-domain and out-of-domain settings, bigger corpus yields better results.

\subsection{Effects of LLM Capabilities}
\begin{figure}[t!]
  \centering
  \begin{subfigure}{0.49\linewidth}
    \centering
    \includegraphics[width=\linewidth]{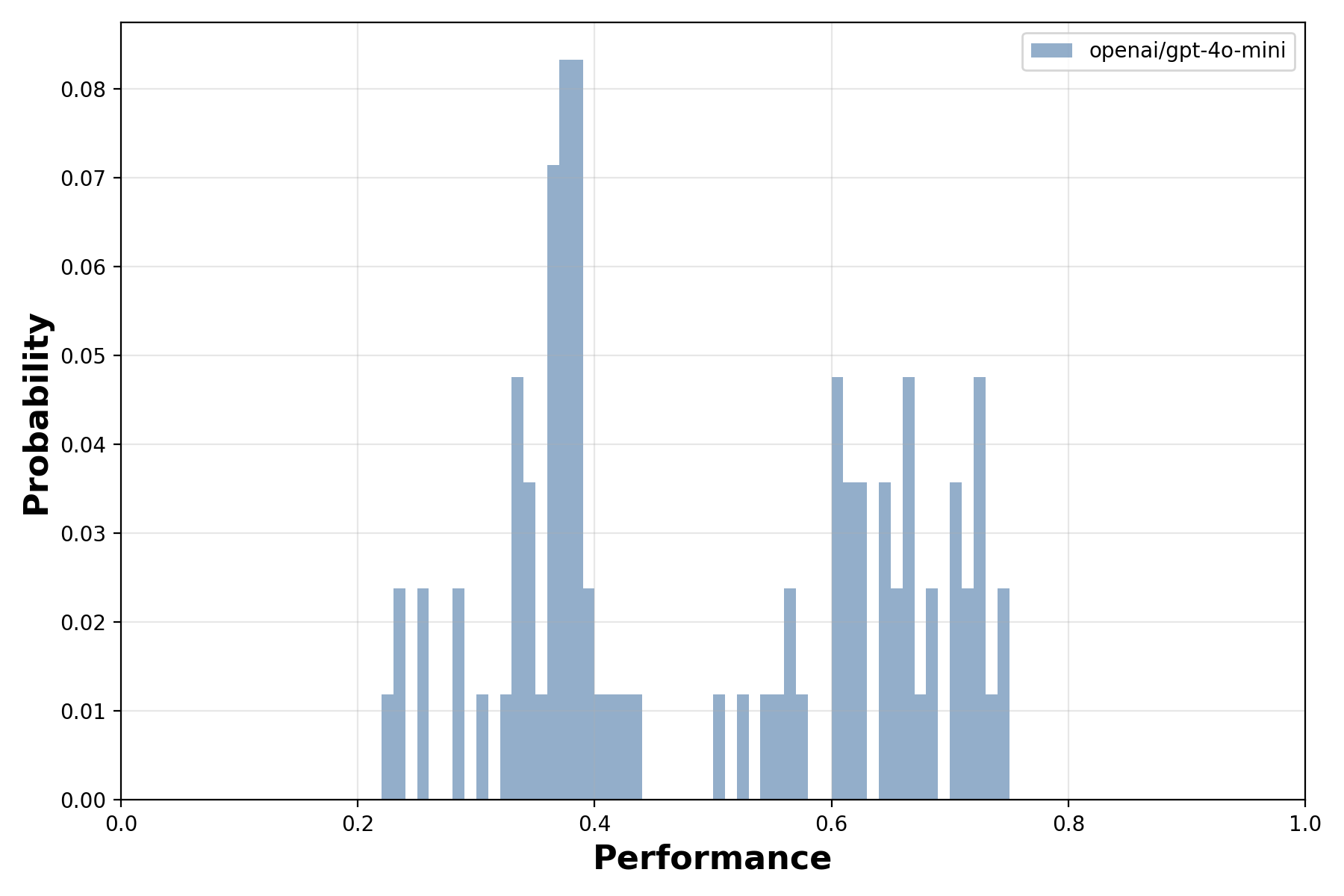}
    \includegraphics[width=\linewidth]{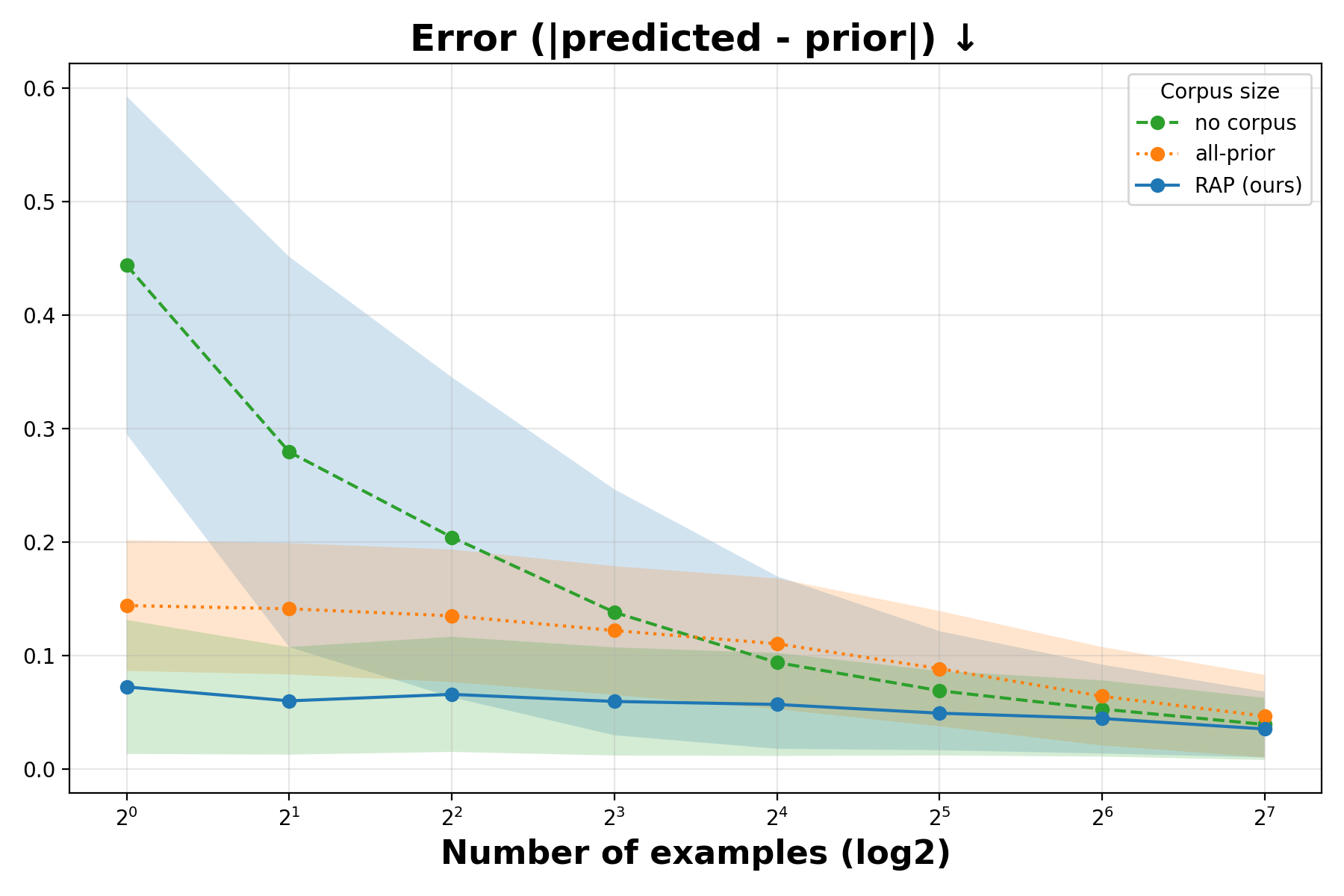}
    \includegraphics[width=\linewidth]{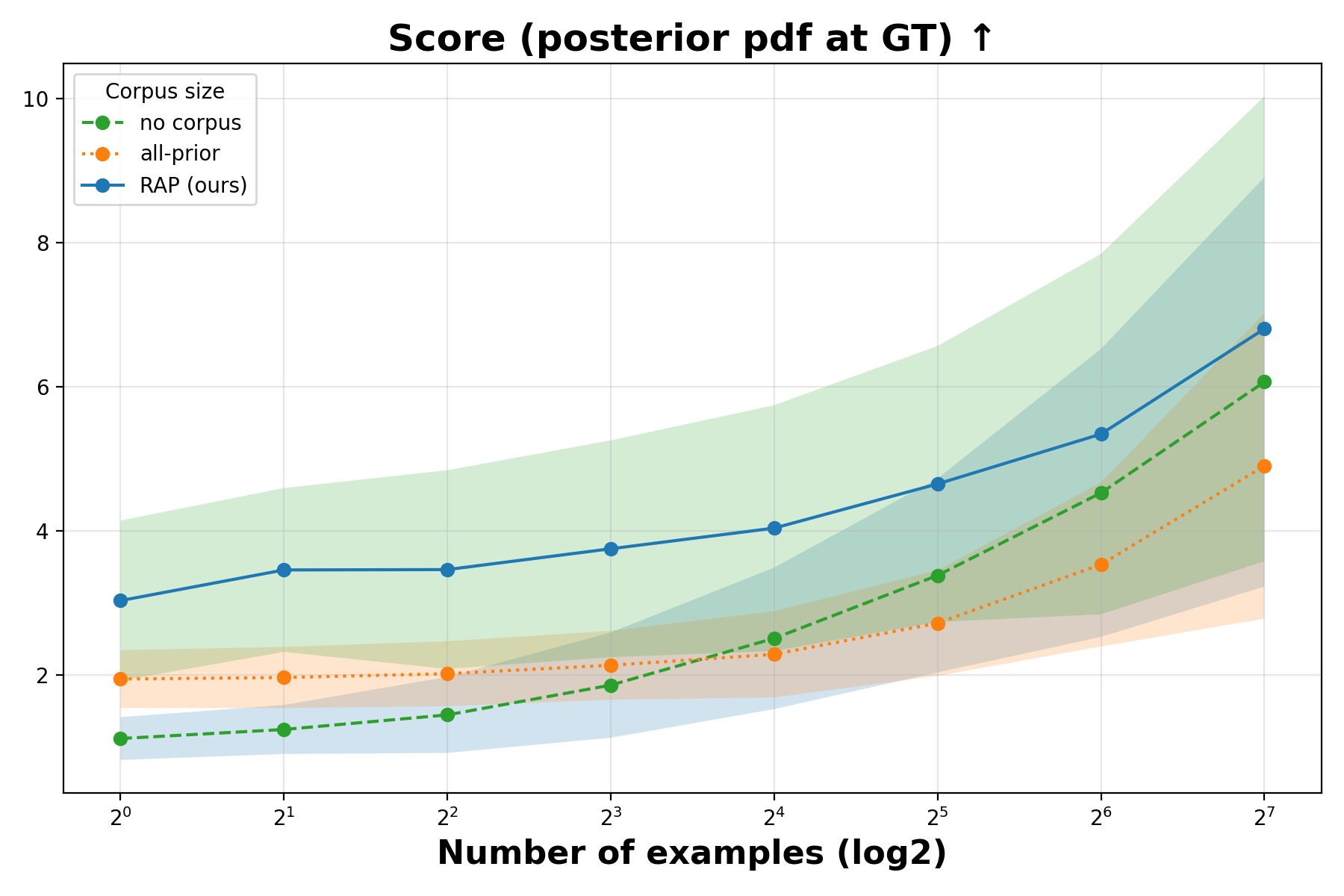}
    \caption{4o-mini}
  \end{subfigure}
  \hfill
  \begin{subfigure}{0.49\linewidth}
    \centering
    \includegraphics[width=\linewidth]{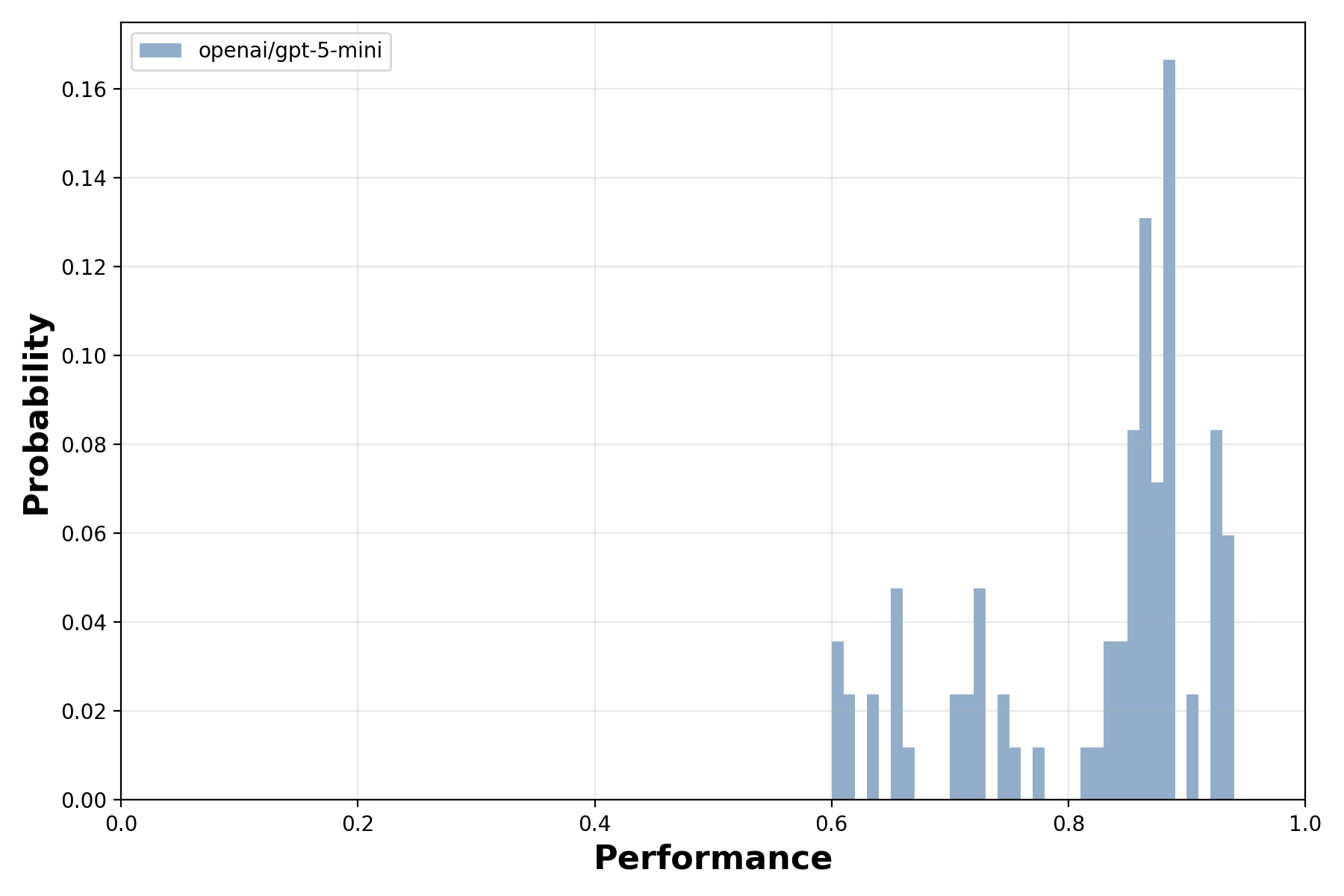}
    \includegraphics[width=\linewidth]{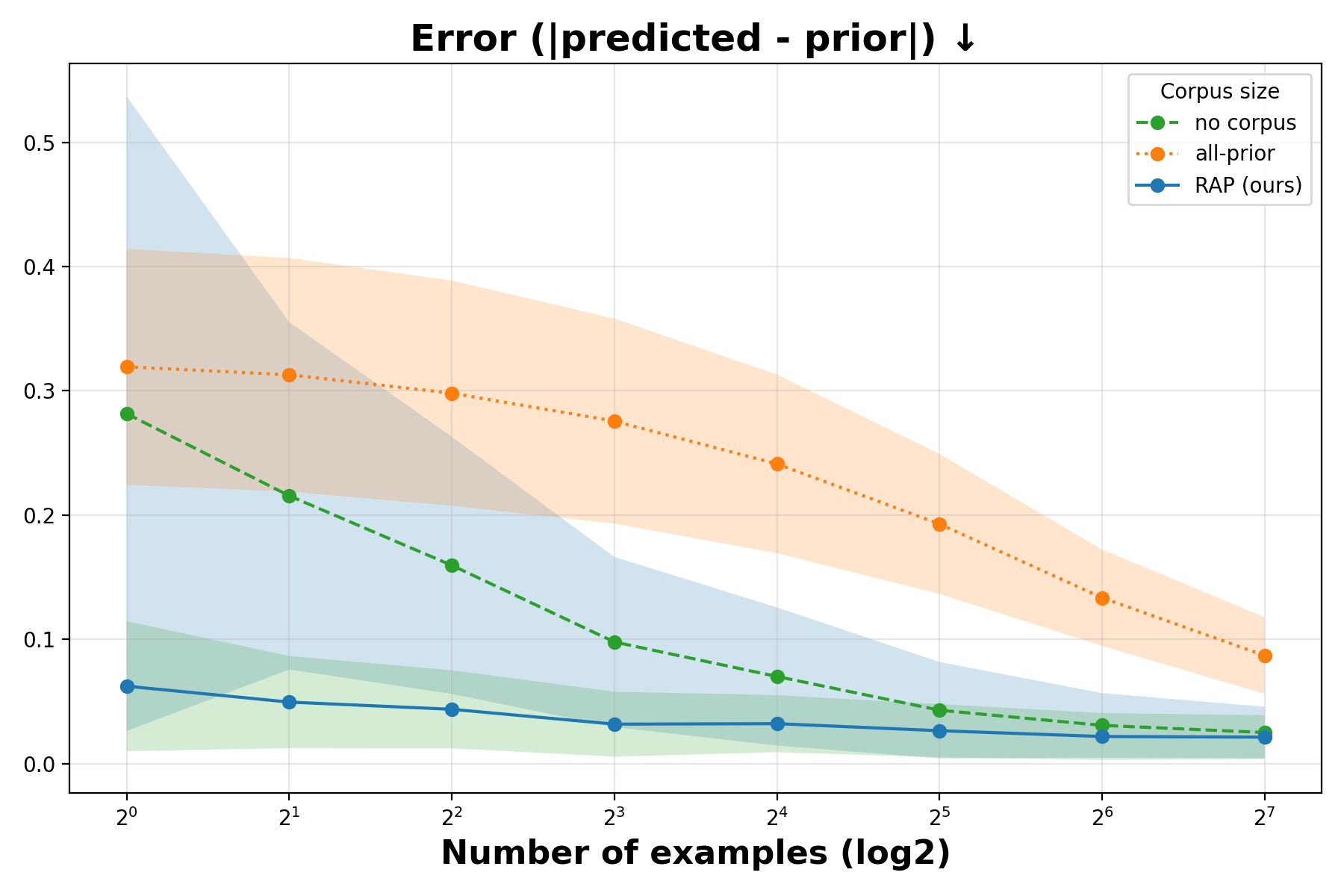}
    \includegraphics[width=\linewidth]{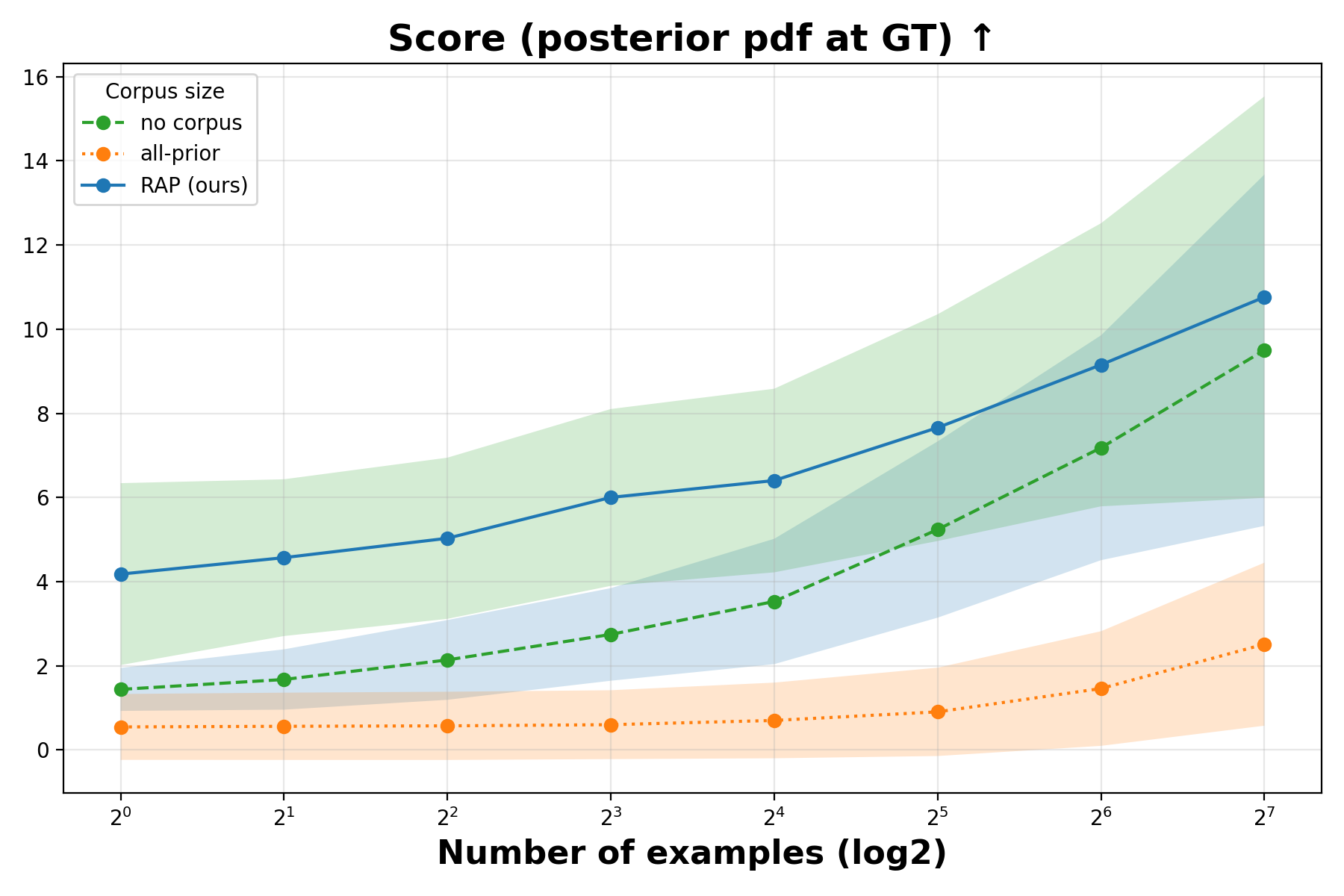}
    \caption{5-mini}
  \end{subfigure}
  \caption{First row: the performance prior for \PP{} if using a weaker model (GPT-4o-mini) versus a stronger model (GPT-5-mini). 
  Stronger model has a prior that's more similar to the prior for symbolic programs. 
  Consequently, the \textbf{no corpus} baseline becomes relatively stronger when used on stronger models.
  RAP still performs best in all settings.
  }
  \label{fig:4omini-vs-5mini}
\end{figure}

In our experiments, we find that different language model settings substantially influence prediction performance. We investigate this phenomenon, with results shown in~\Cref{fig:4omini-vs-5mini}
. 
We find that RAP performs comparatively better for weaker LLMs, where the performance prior is more diffused, and worse for stronger LLMs, whose performance prior is similar to that of \SP{}. 
\clearpage
\subsection{Effects of Text Embedding}
\begin{figure}
    \centering
    \includegraphics[width=0.89\linewidth]{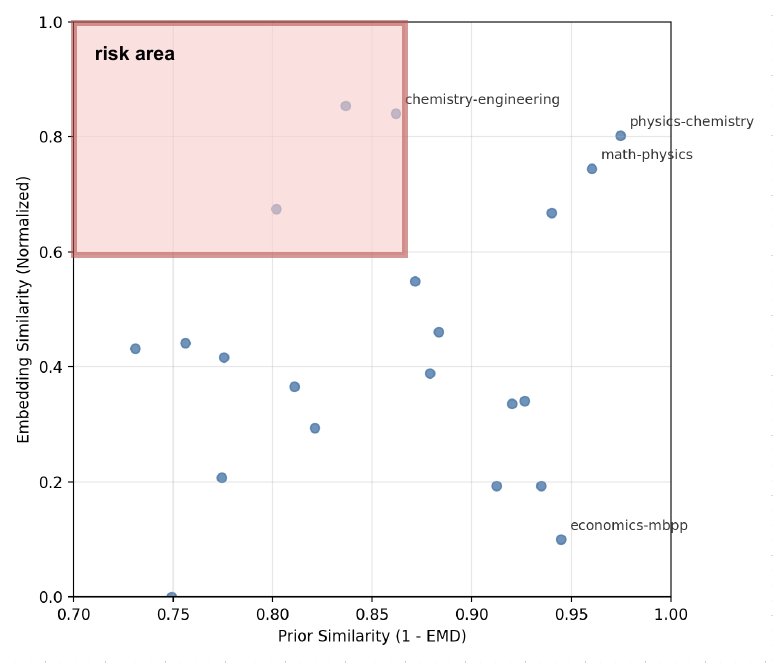}
    \caption{Relationship between textual similarity (of pairs of domains) and performance prior similarity (of pairs of domains).  
Retrieval based on textual similarity does not guarantee similarity in performance prior, for instance, the pair chemistry-engineering. 
The upper-left corner is the \emph{risk area} where RAP might retrieve tasks which induces a prior dis-similar to the target domain.
RAP mitigate this issue with prior strength normalization.}
    \label{fig:risk_area}
\end{figure}

As RAP will always retrieve tasks that are textually similar, it is good to understand if textual similarity implies performance prior similarity.
\Cref{fig:risk_area} shows the correlation between textual similarity and performance prior similarity across different domains.
As we can see, there exists certain pairs of domains (e.g. chemistry-engineering) which are very similar in the textual space (y-axis) but are dissimilar in the prior (x-axis). 
This is the ``risk area'' of our algorithm, where the retrieved tasks do not approximate the prior on the target domain.
RAP mitigate this issue with prior strength normalization.

\end{document}